\documentclass[lettersize,journal]{IEEEtran}
\usepackage{amsmath,amsfonts}
\usepackage{array}
\usepackage[caption=false,font=normalsize,labelfont=sf,textfont=sf]{subfig}
\usepackage{textcomp}
\usepackage{stfloats}
\usepackage{url}
\usepackage{verbatim}
\usepackage{graphicx}
\usepackage{cite}
\usepackage{multirow}
\usepackage[figuresright]{rotating}
\usepackage{amsmath}
\usepackage{colortbl}
\usepackage{color}
\usepackage{caption}
\usepackage{wrapfig}
\usepackage{makecell}
\usepackage{tabularx}
\usepackage[hidelinks]{hyperref}
\usepackage[ruled,vlined,linesnumbered]{algorithm2e}
\SetKw{KwBy}{by}

\usepackage{xcolor}
\definecolor{MyBlue}{HTML}{11144C}
\definecolor{MyGreen}{HTML}{3A9679}
\definecolor{MyYellow}{HTML}{FABC60}
\definecolor{MyRed}{HTML}{E16262}
\definecolor{Hyper}{HTML}{B63618}
\definecolor{MyPurple}{HTML}{6F69AC}
\definecolor{LinkPink}{HTML}{D81B60}

\definecolor{newgreen}{rgb}{0, 0.6, 0.2}

\usepackage{amsmath,amsfonts,bm}

\def\eqref#1{equation~\ref{#1}}

\def\1{\bm{1}}

\def\rmP{{\mathbf{P}}}
\def\rmQ{{\mathbf{Q}}}

\DeclareMathAlphabet{\mathsfit}{\encodingdefault}{\sfdefault}{m}{sl}
\SetMathAlphabet{\mathsfit}{bold}{\encodingdefault}{\sfdefault}{bx}{n}

\def\gA{{\mathcal{A}}}

\def\gC{{\mathcal{C}}}

\def\gM{{\mathcal{M}}}

\def\gS{{\mathcal{S}}}

\def\sZ{{\mathbb{Z}}}

\begin{document}

\title{MesonGS++: Post-training Compression of 3D Gaussian Splatting with Hyperparameter Searching}
\author{Shuzhao Xie$^{\star}$,
~Junchen Ge$^{\star}$,
~Weixiang Zhang, 
~Jiahang Liu, 
~Chen Tang,
~Yunpeng Bai,
~Shijia Ge, \\
~Jingyan Jiang,
~Yuzhi Huang,
~Fengnian Yang,
~Cong Zhang,
~Xiaoyi Fan,
~Zhi Wang$^{\dagger}$,~\IEEEmembership{Senior Member,~IEEE}
\thanks{
  $^{\star}$Equal contribution.
  $^{\dagger}$Zhi Wang is the corresponding author.
}
\thanks{
  Shuzhao Xie, Junchen Ge, Shijia Ge, Fengnian Yang, and Zhi Wang are with Shenzhen International Graduate School,
  Tsinghua University, Shenzhen, China.
  Weixiang Zhang is with The Hong Kong University of Science and Technology,
  Hong Kong SAR, China.
  Jiahang Liu is with Harbin Institute of Technology, Shenzhen, China.
  Chen Tang is with The Chinese University of Hong Kong, Hong Kong SAR, China.
  Yunpeng Bai is with The University of Texas at Austin, USA.
  Jingyan Jiang is with Shenzhen Technology University, China.
  Yuzhi Huang is with Xiamen University, China.
  Cong Zhang is with Simon Fraser University, Canada.
  Xiaoyi Fan is with Jiangxing Intelligence Inc., China.
}
}

\markboth{Journal of \LaTeX\ Class Files,~Vol.~14, No.~8, August~2021}%
{Shell \MakeLowercase{\textit{et al.}}: A Sample Article Using IEEEtran.cls for IEEE Journals}

\IEEEpubid{0000--0000/00\$00.00~\copyright~2021 IEEE}

\maketitle

\begin{abstract}
    3D Gaussian Splatting (3DGS) achieves high-quality novel view synthesis with real-time rendering, but its storage cost remains prohibitive for practical deployment. Existing post-training compression methods still rely on many coupled hyperparameters across pruning, transformation, quantization, and entropy coding, making it difficult to control the final compressed size and fully exploit the rate-distortion trade-off. We propose \emph{MesonGS++}, a size-aware post-training codec for 3D Gaussian compression. On the codec side, MesonGS++ combines joint importance-based pruning, octree geometry coding, attribute transformation, selective vector quantization for higher-degree spherical harmonics, and group-wise mixed-precision quantization with entropy coding. On the configuration side, it treats the reserve ratio and bit-width allocation as the dominant rate-distortion knobs and jointly optimizes them under a target storage budget via discrete sampling and 0--1 integer linear programming. We further propose a linear size estimator and a CUDA parallel quantization operator to accelerate the hyperparameter searching process. Extensive experiments show that MesonGS++ achieves over 34$\times$ compression while preserving rendering fidelity, outperforming state-of-the-art post-training methods and accurately meeting target size budgets. Remarkably, without any training, MesonGS++ can even surpass the PSNR of vanilla 3DGS at a 20$\times$ compression rate on the \textit{Stump} scene. Our code is available at \href{https://github.com/mmlab-sigs/mesongs_plus}{\textcolor{LinkPink}{here}}.
\end{abstract}


\section{Introduction}

\IEEEPARstart{N}{ovel} view synthesis is a fundamental task in 3D vision and 
has significant applications in virtual reality~\cite{luo2025vrdoh,jiang2024vr,yao2025sd,Zhang_2025_CVPR,sympowerinr,zhang2025understanding,gu2024dragscene}, autonomous driving~\cite{zhou2024drivinggaussian,yan2024street,Wang_2025_CVPR}, and robotics~\cite{zhao2025real2edit2real,fan2025twinaligner,gu2025igen}.
This task involves using a collection of images captured from different viewpoints along with their corresponding camera poses, 
with the objective of generating highly realistic images from arbitrary viewpoints.
By reparameterizing the point with a 3D Gaussian function in the point cloud,
3D Gaussian Splatting (3D-GS) \cite{kerbl20233d} shows 
excellent quality and real-time rendering speed in this task.
A Gaussian point consists of a 3D coordinate, spherical harmonics (SH) coefficients to represent its color, an opacity parameter, a scale vector, and a rotation quaternion. 
3D-GS utilizes scale vectors and rotation quaternions to characterize the covariance matrix of the 3D Gaussian function. The coordinates of Gaussians are usually referred as \textit{geometry} and the other parameters of Gaussians are referred as \textit{attributes}.
Despite the efficiency of 3D-GS, the sheer volume of Gaussians and the multi-channel attributes within each Gaussian result in a considerable file size. Notably, $5.27 \times 10^6$ Gaussians are required to represent the \textit{bicycle} scene in the Mip-NeRF 360 dataset \cite{barron2022mip}, occupying 1.3 GB of storage under 32-bit float precision. This sizable file poses challenges in the transmission and storage.
Hence, it is essential to design a tailored codec for 3D Gaussians.

\IEEEpubidadjcol

Due to the heterogeneous attributes and intricate rendering pipeline of 3D Gaussian Splatting, designing an efficient compression framework for 3D Gaussians is highly nontrivial. Although a growing body of 3DGS compression methods has been proposed, many of them~\cite{niedermayr2023compressed,fan2023lightgaussian,navaneet2023compact3d,morgenstern2023compact,jc2023gsplat,ap2023gssmall,ap2023gssmaller,pcgs2025,zhan2025cat,xie2024mesongs,xie2024sizegs,ramlot2026potr,liu2025splatwizard,zhang2025gaussianspa,wu2024implicitgs,liu2024compgs,hac++2025,hac2024,girish2023eagles,wang2024end,wang2024contextgs,papantonakis2024i3d,shin2025locality,cao2024lightweight,zhang2024lp,hanson2025pup,hanson2025speedy,niemeyer2025radsplat,fang2024mini,fang2024mini2,chen2025megs,lee2026safeguardgs,huang2025entropygs,liu2025efficientgs,tian2025flexgaussian} operate during training and obtain compact models by modifying the optimization procedure itself. In many practical scenarios, however, one instead wishes to re-encode an already trained 3DGS model afterward, much like conventional image or video compression. This motivates post-training 3DGS compression~\cite{xie2024mesongs,xie2024sizegs,tian2025flexgaussian,fcgs2024,ramlot2026potr}, which starts from an existing representation and compresses it without retraining the full model from scratch.

The central challenge of post-training compression is not only \emph{how} to compress, but also \emph{how to configure} the compression pipeline. A practical post-training codec~\cite{xie2024mesongs,xie2024sizegs,tian2025flexgaussian,ramlot2026potr} typically involves multiple stages, such as pruning, transformation, quantization, and entropy coding, each introducing its own hyperparameters. These hyperparameters are strongly coupled: pruning changes the attribute distribution seen by the downstream transform and quantizer; transformation affects entropy and quantization sensitivity; and bit-width allocation determines both quality loss and final file size. As a result, existing post-training methods often rely on manual tuning or locally designed heuristics, which provide limited global control over the final compressed size and frequently force overly coarse quantization configurations. This issue becomes particularly problematic when the target file size is a hard constraint, as in transmission and storage-limited applications.

To address this problem, we propose \emph{MesonGS++}, a size-aware post-training codec for 3D Gaussian compression. The key idea is to co-design the compression operators and the hyperparameter configuration mechanism, so that the codec is not only effective in reducing redundancy, but also structured for explicit size-constrained compression. On the codec side, MesonGS++ first prunes insignificant Gaussians using a joint importance metric that combines view-dependent rendering contribution with a view-independent volume prior, making pruning robust in both bounded and unbounded scenes. It then reduces attribute redundancy through two transformations that better suit entropy coding, namely replacing rotation quaternions with Euler angles and applying region adaptive hierarchical transform (RAHT)~\cite{de2016compression} to key attributes to expose low-entropy AC coefficients. For higher-degree spherical harmonics, we retain a subset directly and compress the remainder with vector quantization~\cite{equitz1989new} to better preserve rendering quality. Geometry is compressed by octree voxelization and encoded by G-PCC~\cite{tmc13}. The transformed attributes are quantized by group-wise mixed precision and encoded by torchac~\cite{mentzer2019practical} so that different parameter groups can use different bit-widths, avoiding the quality bottleneck of coarse uniform quantization.

On the configuration side, MesonGS++ treats the reserve ratio $\tau$ and group-wise mixed-precision bit allocation as the dominant rate-distortion knobs, and jointly optimizes them under a prescribed storage budget. Specifically, we formulate this configuration problem as a mixed integer nonlinear programming problem, and solve it efficiently through discrete sampling of $\tau$ and a 0-1 ILP for bit allocation. We further introduce a linear size estimator, use quantization loss as a proxy for visual degradation, and parallelize group-wise quantization to accelerate the search. When finetuning is allowed, a piecewise finetuning strategy is used to restore reconstruction quality. In this way, MesonGS++ unifies high-quality post-training compression and accurate target-size control within a single framework.

Our contributions can be summarized as follows:
\begin{itemize}
    \item We propose \emph{MesonGS++}, a size-aware post-training codec for 3D Gaussian compression that unifies joint importance-based pruning, octree geometry coding, attribute transformation, selective vector quantization for higher-degree spherical harmonics, group-wise mixed-precision quantization, and entropy coding in a single framework.
    \item We identify hyperparameter configuration as a central bottleneck of post-training 3DGS compression, and formulate the dominant configuration variables, namely the reserve ratio and group-wise bit-width allocation, as a budget-constrained optimization problem. We solve it efficiently through discrete sampling of $\tau$, 0-1 ILP for mixed-precision bit allocation. A linear size estimator and a parallel group-wise quantization scheme are designed for acceleration.
    \item Extensive experiments demonstrate that MesonGS++ achieves over 34$\times$ compression while maintaining fidelity, surpassing state-of-the-art post-training methods in rate-distortion performance, with accurate target-size control and favorable speed-quality trade-offs. Remarkably, without any training, MesonGS++ can even surpass the PSNR of vanilla 3DGS at a 20$\times$ compression rate. The resulting compression configuration also exhibits transferability to other compressed Gaussian representations.
\end{itemize}

\section{Preliminary}
3D-GS \cite{kerbl20233d} is an explicit 3D scene representation in the
form of point clouds, utilizing Gaussians to model the
scene. Each Gaussian is characterized by a covariance matrix
$\mathbf{\Sigma}$ and a center point $\mathbf{X}$, which is referred to as the mean value of the Gaussian:
\begin{equation}
    \label{eq:gs_rep}
    G(x) = e^{-\frac{1}{2}\mathbf{X}^\top \mathbf{\Sigma}^{-1}\mathbf{X}}.
\end{equation}
To maintain the positive definiteness of the covariance matrix $\mathbf{\Sigma}$, 
3D-GS decomposes $\mathbf{\Sigma}$ into a scaling matrix $\mathbf{S} = {\rm diag}(\mathbf{s}), \mathbf{s} \in \mathbb{R}^3$ and a rotation matrix $\mathbf{R}$: $\mathbf{\Sigma} = \mathbf{R}\mathbf{S}\mathbf{S}^\top \mathbf{R}^\top$.
The rotation matrix $\mathbf{R}$ is parameterized by a rotation quaternion $\mathbf{q} \in \mathbb{R}^4$. The backpropagation process is illustrated in \cite{kerbl20233d}.

When rendering novel views, the technique of splatting
\cite{ewa_volume_splatting,yifan2019differentiable} is employed for the Gaussians 
within the camera planes. As introduced by \cite{zwicker2001surface}, 
using a viewing transform denoted as $\mathbf{W}$ and 
the Jacobian $\mathbf{J}$ of the affine approximation
of the projective transformation, the covariance
matrix $\mathbf{\Sigma}'$ in camera coordinates system can be computed by $\mathbf{\Sigma}' = \mathbf{J} \mathbf{W} \mathbf{\Sigma} \mathbf{W}^\top \mathbf{J}^\top$.

In summary, each element of 3D Gaussians has the following parameters:
(1) a 3D center $\mathbf{\mu} \in \mathbb{R}^3$; (2) a rotation quaternion $\mathbf{q} \in \mathbb{R}^4$;
(3) a scale vector $\mathbf{s} \in \mathbb{R}^3$; 
(4) a color feature defined by spherical harmonics coefficients $\mathbf{SH} \in \mathbb{R}^d$, with $d = 3(f+1)^2$, where
$f$ is the harmonics degree; 
and (5) an opacity logit $o \in \mathbb{R}$. 
Specifically, for each pixel, the color and opacity of all the Gaussians are computed using Eq.~\ref{eq:gs_rep}. 
The blending of $N$ ordered points that overlap the pixel is given by:
\begin{equation}
    \label{eq:alpha_comp}
    C = \sum_{i\in N}{c_i \alpha_i \prod_{j=1}^{i-1}(1 - \alpha_j)}.
\end{equation}
Here, $c_i$ and $\alpha_i$ represent the density and color of this point
computed by a Gaussian with covariance $\mathbf{\Sigma}$ multiplied by
an optimizable per-point opacity and SH color coefficients.

\section{Methodology}
\begin{figure}[!t]
    \centering
    \subfloat[\textit{Bounded scenes.}\label{fig:syn_cdf}]{%
        \includegraphics[width=0.48\linewidth]{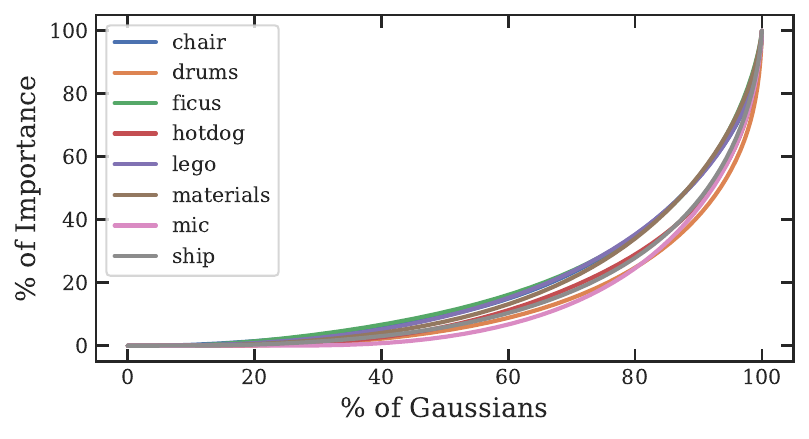}%
    }
    \hfill
    \subfloat[\textit{Unbounded scenes.}\label{fig:ub_cdf}]{%
        \includegraphics[width=0.48\linewidth]{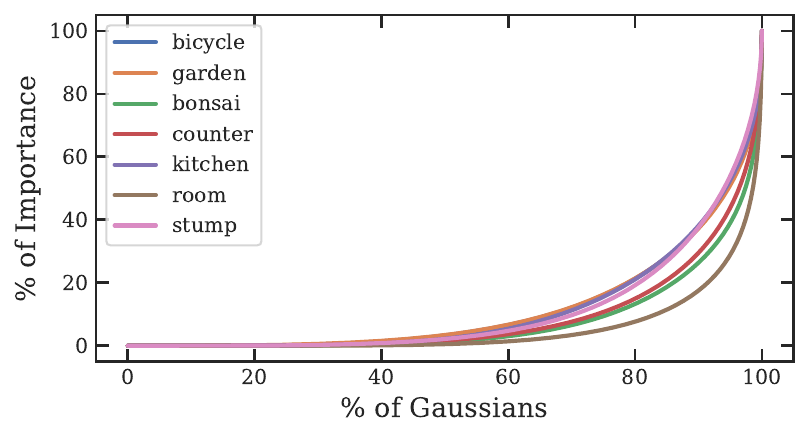}%
    }
    \caption{\textbf{Quantile-quantile curve.} $\mathbf{x}$\protect\% of the least important Gaussians
    contribute to $\mathbf{y}$\protect\% of total importance. For both kinds of scenes,
    40\protect\% of the Gaussians contribute over 80\protect\% of the importance. The importance refers to the contribution to the final rendering results.}
    \label{fig:cdf}
\end{figure}

\begin{figure*}[t]
    \centering
     \includegraphics[width=\linewidth]{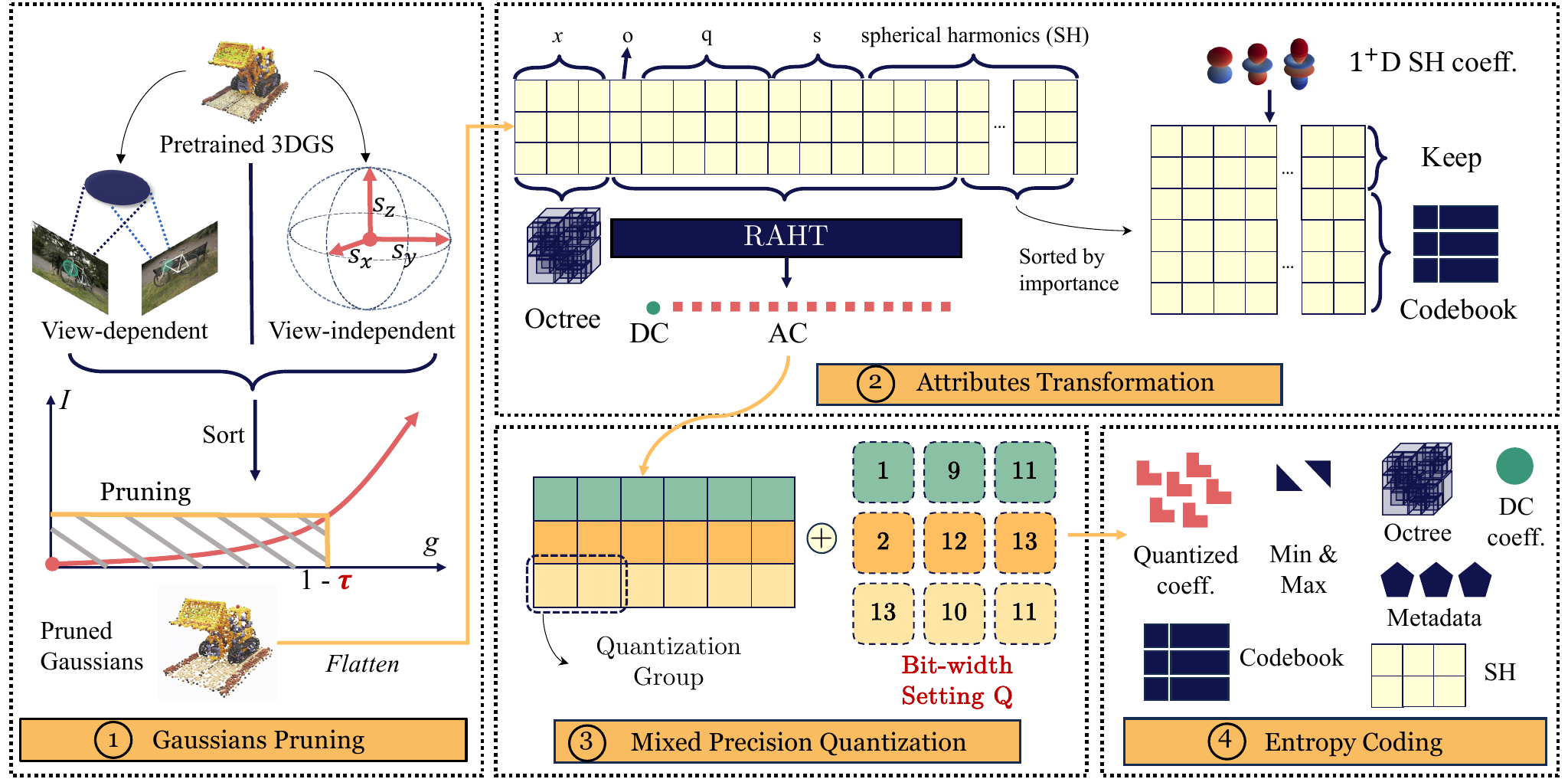}
     \caption{\textbf{Overview of the proposed compression pipeline.} \textcircled{\raisebox{-0.5pt}{1}} We first prune insignificant Gaussians using a joint importance metric that combines view-dependent and view-independent cues. \textcircled{\raisebox{-0.5pt}{2}} We then compress geometry by building an octree to obtain voxelized coordinates, and apply RAHT to opacity, quaternion, and degree-0 spherical harmonics. For higher-degree SH coefficients, we retain a subset directly and compress the remainder with vector quantization to preserve rendering quality. RAHT is not applied to scales when the quantization bit-width is 8. \textcircled{\raisebox{-0.5pt}{3}} Next, we perform group-wise quantization on the RAHT-transformed AC coefficients, allowing different parameter groups to use different bit-widths. \textcircled{\raisebox{-0.5pt}{4}} Finally, all components are encoded by our entropy coding scheme. Since the pipeline involves numerous interdependent hyperparameters whose joint optimization is critical to achieving the best rate-distortion performance, we further introduce an automated hyperparameter configuration algorithm; see Fig.~\ref{fig:hsearch}.}
    \label{fig:overview}
    \vspace{-0.5cm}
\end{figure*}

In this section, we first present the proposed post-training compression pipeline for 3D Gaussian Splatting, as illustrated in Fig.~\ref{fig:overview}. The pipeline comprises five stages: (1) pruning insignificant Gaussians using a joint importance metric that combines view-dependent and view-independent cues; (2) compressing the geometry via an octree structure over voxelized coordinates; (3) transforming important attributes with the Region Adaptive Hierarchical Transform (RAHT) and applying vector quantization to higher-degree spherical harmonics (SH) coefficients; (4) applying group-wise quantization to the RAHT-transformed AC coefficients, allowing different attribute groups to use different bit-widths; and (5) encoding all components via entropy coding. After that, to find the optimal hyperparameter configuration, specifically the reserve ratio and quantization bit-widths, under a given size constraint, we formulate the search as a Mixed Integer Nonlinear Programming (MINLP) problem and propose an efficient hierarchical solver, as illustrated in Fig.~\ref{fig:hsearch}.

\subsection{Gaussians Pruning}
As 3D-GS has a huge number of points, pruning unimportant Gaussians is a necessary step.
We first define the importance score of each Gaussian. Here the importance score refers to the contribution to the final rendering results.
For a Gaussian $g$, we define its importance score $I_g$ as the product of the view-dependent importance score $I_d$ and the view-independent importance score $I_i$: $I_g = I_d I_i$.
Based on Eq.~\ref{eq:alpha_comp},
we define view-dependent importance score $I_d$ as: 
\begin{equation}
    I_d = \sum_{p \in \mathcal{P}} \alpha_i \prod_{j=1}^{i-1}(1 - \alpha_j).
    \label{eq:view_dep}
\end{equation}
Here $\mathcal{P}$ is the pixel set that are overlapped by the Gaussian $g$,
and $i$ is the rank of Gaussian $g$ in a set of Gaussians that overlap with the pixel $p$.
In contrast to VQRF \cite{li2023compressing}, 
where the importance score is the mean value of corresponding sample points, our method allows for the direct recording of the importance score throughout the testing phase.
LightGaussian \cite{fan2023lightgaussian} uses the opacity ($\alpha_i$ in Eq.~\ref{eq:view_dep}) as the view-dependent score. They have not considered the masking caused by other Gaussians. 
The usage of backward gradients as importance scores in C3DGS \cite{niedermayr2023compressed} is not ideal for 3D-GS that is not well-learned. Typically, Gaussians that are not well-learned exhibit larger gradients. However, these poorly-learned Gaussians may not be the significant ones, making the pruning strategy of C3DGS ineffective in inadequately learned 3D Gaussians.

The view-independent score $I_i$ is given by $I_i = (V_{\rm norm})^{\beta}$.
Here the volume $V$ is the product of the scale vector.
To obtain $V_{\rm norm}$, we normalize the $V$ by the 90\% largest
of all sorted Gaussians and clip the range between 0 and 1,
$\beta$ is the hyperparameter to control the size of $I_i$.

In Fig.~\ref{fig:cdf},
we sort the importance score and visualize its cumulative distribution function (CDF). We notice that 40\% of the Gaussians contain over 80\% of the importance. Hence, we use a importance threshold $\tau$ to set the reserve ratio of Gaussians, which means that we cut the percent of $1 - \tau$ of the least important Gaussians.

\subsection{Geometry Compression}
After pruning, we compress the 3D positions with the octree structure. An octree recursively divides occupied voxels into 8 subvoxels until the required resolution is reached. The occupancy symbol is composed of 8 bits (1 to 255 in decimal), where each bit indicates the occupancy status of the corresponding subvoxel. We use the depth $d$ to control the size of the octree. When multiple Gaussians existing within a voxel, we average the corresponding attributes for deduplication.

\subsection{Attribute Transformation}
We categorize attributes into important attributes and unimportant attributes. Important attributes include opacity, scales, rotations/Euler angles, and 0D-SH coefficients. Here, 0D-SH coefficient refers to SH coefficients in degree 0. The unimportant attributes refer to the SH coefficients in degrees greater than 0. 

\vspace{1mm}\noindent\textbf{Replacement.}
We replace the rotation quaternion (4 numbers) with the corresponding Euler angles (3 numbers).
The Euler angles are three angles to describe the orientation of a rigid body with respect to a fixed coordinate system. This replacement can reduce a number of storage.

Specifically, for a quaternion $\mathbf{q} = [w, x, y, z] \in \mathbb{R}^4$, 
we calculate the euler angle $\mathbf{e} = [\phi, \theta, \psi] \in \mathbb{R}^3$ with:
\begin{equation}\scriptstyle
\left[ 
    \begin{array}{c}
        {\rm atan2}(2(wx + yz), 1 - 2(x^2 + y^2)) \\
        -\frac{\pi}{2} + 2{\rm atan2}(\sqrt{1 + 2(wy - xz)}, \sqrt{1 - 2(wy - xz)}) \\
        {\rm atan2}(2(wz+xy), 1 - 2(y^2 + z^2))
    \end{array}
\right].
\end{equation}
During decoding, we directly build the rotation matrix $\mathbf{R}$ from the Euler angles by:
\begin{equation}\scriptstyle
\left[ 
\begin{array}{ccc}
    {\rm C}_{\theta}{\rm C}_{\psi} & -{\rm C}_{\phi}{\rm S}_{\psi} + {\rm S}_{\phi}{\rm S}_{\theta}{\rm C}_{\psi} & {\rm S}_{\phi}{\rm S}_{\psi} + {\rm C}_{\phi}{\rm S}_{\theta}{\rm C}_{\psi} \\
    {\rm C}_{\theta}{\rm S}_{\psi} & {\rm C}_{\phi}{\rm C}_{\psi} + {\rm S}_{\phi}{\rm S}_{\theta}{\rm S}_{\psi} & -{\rm S}_{\phi}{\rm C}_{\psi} + {\rm C}_{\phi}{\rm S}_{\theta}{\rm S}_{\psi} \\
    -{\rm S}_{\theta} & {\rm S}_{\phi}{\rm C}_{\theta} & {\rm C}_{\phi}{\rm C}_{\theta} \\
\end{array}
\right].
\end{equation}
Here ${\rm S}_{\theta}$ and ${\rm C}_{\theta}$ represents sine and cosine of $\theta$. Similarly, $\phi$ and $\psi$ follow the same notation.

As the covariance matrix is symmetry, replacing the scales and rotation quaternions (7 numbers) with the upper triangular part (6 numbers) of the covariance matrix seems to be an alternative.
This strategy can also reduce the storage of one number.
Our contemporary, C3DGS\cite{niedermayr2023compressed}, proposes a similar covariance-based replacement strategy. 
However, the subsequent quantization steps will 
cause a large number of covariance matrices 
to become indefinite, resulting in degraded rendering outcomes.
In contrast, using Euler angles can ensure 
that the positive definiteness of covariance is not compromised.
The comparison results are illustrated in Fig.~\ref{fig:euler_vs_cov}.

\begin{figure*}[t]
    \centering
    \begin{minipage}[b]{0.31\textwidth}
        \centering
        \includegraphics[width=0.99\linewidth]{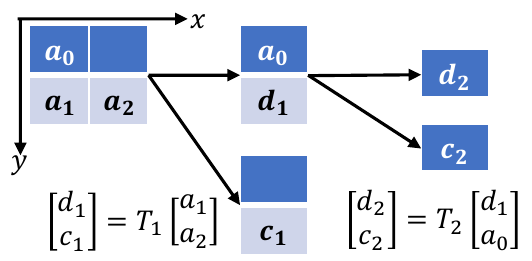}
        \caption{\textbf{2D example of RAHT.}}
        \label{fig:raht}
    \end{minipage}
    \begin{minipage}[b]{0.68\textwidth}
        \centering
        \includegraphics[width=0.99\linewidth]{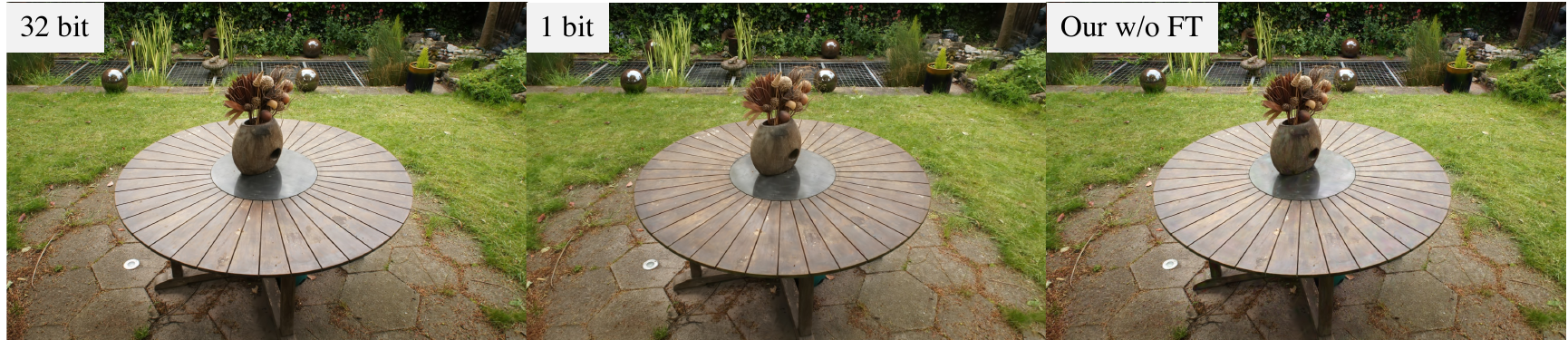}
        \caption{\textbf{Effect of $1^+$D spherical harmonics coefficients.} ``FT'' refers to finetune.}
        \label{fig:cluster}
    \end{minipage}
    \vspace{-0.5cm}
\end{figure*}

\vspace{1mm}\noindent\textbf{Region Adaptive Hierarchical Transform.}
To reduce signal redundancy in important attributes, 
we apply RAHT~\cite{de2016compression} 
to the remained attributes.
RAHT involves taking voxelized coordinates from octree
and converting the corresponding attributes into transformed coefficients. 
Each channel of the transformed coefficients consists of a direct current (DC) coefficient
and serveral alternating current (AC) coefficients.
The low entropy of the AC coefficients enables the subsequent entropy coding procedure to achieve larger compression rate. 
Here we briefly introduce the RAHT through a 2D example.
Fig.~\ref{fig:raht} shows how to apply RAHT to $a_0$, $a_1$, and $a_2$.
First, RAHT merges the coefficients along the $x$ axis with transform $T_1$:
\begin{equation}
    T_1 = \frac{1}{\sqrt{w_1 + w_2}} 
    \left[
        \begin{array}{cc}
        \sqrt{w_1} & \sqrt{w_2} \\
        -\sqrt{w_2} & \sqrt{w_1} \\
        \end{array}
    \right],
\end{equation}
where $w_i$ is the number of Gaussians in $a_i$ (1 for leaf nodes; the sum of child weights otherwise).
Applying $T_1$ to $a_1$ and $a_2$ yields DC coefficient $d_1$ and AC coefficient $c_1$; $d_1$ is further transformed with $a_0$, while AC coefficients are retained for encoding. Unmatched coefficients (e.g., $a_0$) pass to the next layer, ultimately producing one DC coefficient $d_2$ and two AC coefficients $c_1$, $c_2$. Weight coefficients are recovered from the octree during decoding.

\vspace{1mm}\noindent\textbf{Clustering $1^+$D SH Coefficients.}
The size of spherical harmonics (SH) coefficients takes up 85.7\% in a 3D Gaussians file. In the middle of Fig.~\ref{fig:cluster}, after applying 1-bit quantization to the $1^+$D SH coefficients, only the reflectance is affected, while the overall structure and color remain unchanged. 
Therefore, quantization is not the optimal choice for $1^+$D SH coefficients.
We employ vector quantization to significantly reduce the size of these SH coefficients.
Specifically, we use a codebook and a corresponding index mapping table to 
connect the origin vectors with the vectors in the codebook.
The right side of Fig.~\ref{fig:cluster} shows the final result of our method.
To reduce memory occupation and encoding time, a batched clustering strategy \cite{web_scale_cluster} is used.
Clustering runs for multiple iterations to refine the results. However, replacing every SH vector with a codebook entry inevitably introduces approximation errors, particularly for high-importance Gaussians. To mitigate this, we adopt a budget-first allocation strategy. We first estimate the compressed size of essential components, including octree geometry, VQ indices, base codebook, and quantized RAHT coefficients. Then, we allocate any remaining budget to preserve the original $1^+$D SH coefficients of the top-ranked Gaussians by importance score. These retained vectors are appended to the codebook and encoded jointly.

\subsection{Group-wise Attribute Quantization} After applying RAHT to important attributes, we save the DC coefficient in float, and quantize the AC coefficients.
We use group-wise quantization~\cite{frantar-gptq,2023-emnlp-osplus,MLSYS2024_5edb57c0,tang2022mixed} to prevent significant quality degradation caused by the coarse-grained channel-wise quantization.
Specifically, we first partition a channel of attributes into multiple groups. Then we quantize a \textbf{quantization group} $\mathbf{c}$ with:
\begin{equation}
    \mathbf{c}_q = \lfloor {\rm clamp}(\frac{\mathbf{c}}{S_c} + Z_c, 0, 2^b - 1) \rceil,
\end{equation}
where
\begin{equation}
    S_c = \frac{\max(\mathbf{c}) - \min(\mathbf{c})}{2^{b}},\space\space Z_c = \lfloor 2^b - \frac{\max(\mathbf{c})}{S_c} \rceil.
\end{equation}
Here $b$ refers to the bit-width, $\lfloor \cdot \rceil$ represents the rounding-to-nearest function, and $\mathbf{c}_q$ refers to the quantized attributes. Besides, function ${\rm clamp}(\cdot)$ specifies a range of values. Values below the minimum are set to the minimum. Values above the maximum are set to the maximum. 

Note that we do not apply RAHT to the scale vectors when the quantization bit is 8.
The reason is that the activation function of the scale vector is exponential function, which magnifies the errors caused by transformation and quantization.
We save the minimum and maximum values of all blocks of quantized attributes for decoding.

\subsection{Hyperparameter Searching as MINLP}
\begin{figure*}[t]
    \centering
     \includegraphics[width=\linewidth]{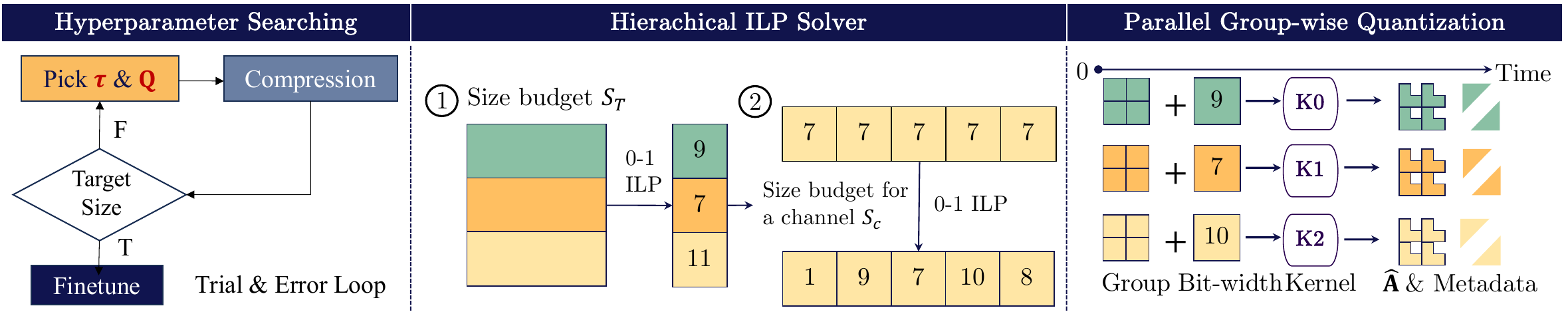}
     \caption{\textbf{Procedure of Hyperparameter Search.} We formulate the hyperparameter search problem as the mixed integer programming, with the optimization goal of reducing quality distortion and constraint by the size. As shown in the circle in the left, each iteration is started by adjusting the \textcolor{Hyper}{reserve ratio $\tau$} and \textcolor{Hyper}{bit-width setting $\rmQ$}. Algo.~\ref{algo:searching} presents our hyperparameter optimization algorithm, which aims to meet the size target while maximizing quality. To accelerate the bit-width setting solver, we use 0--1 ILP to compute the bit-width for each channel, and then construct a 0--1 ILP per channel to solve for the bit-width of each group. Additionally, we implement parallel group-wise quantization to further speed up the process.}
    \label{fig:hsearch}
    \vspace{-0.5cm}
\end{figure*}

\noindent\textbf{Motivation.}~The previous sections introduced efficient compression methods, but these methods also introduce many hyperparameters, such as the reserve ratio $\tau$ and the quantization bit-width of each attribute. For a given size budget, there exists an optimal hyperparameter configuration that maximizes rendering quality, representing the Pareto-optimal trade-off between size and visual quality within this framework. However, these hyperparameters are difficult to tune manually, and a naive configuration is unlikely to reach this optimum. Therefore, a principled search strategy is required to identify the optimal hyperparameter configuration under a specified size constraint. 

\vspace{1mm}\noindent\textbf{Overview.}~Specifically, we first introduce the compression pipeline. Then, we formulate the size-aware 3DGS compression as a mixed integer nonlinear programming (MINLP) problem, aiming to search for the suitable values of the reserve ratio $\tau$ and bit-width settings $\rmQ$. Next, we decouple the MINLP into two subproblems: (1) discrete sampling of $\tau$, and (2) solving for $\rmQ$ using integer linear programming (ILP) with a fixed $\tau$. To fit the ILP solver, we establish an analytical relationship between $\rmQ$ and size. To obtain more accurate solutions, we further model the ILP as a 0--1 ILP problem and refine the size estimation through multiple iterations. To accelerate ILP solver, we adopt the quantization loss to estimate the visual quality degradation and implement a CUDA kernel to accelerate group-wise quantization. Finally, we describe the finetune strategy.

\vspace{1mm}\noindent\textbf{Problem Formulation.} Given the above pipeline, our goal is to search for suitable values of the $\tau$ and $\rmQ$ under a given size constraint, while maximizing visual quality. Hence, we formulate this problem as a mixed integer nonlinear programming (MINLP) model:
\begin{equation}
\begin{aligned}
\label{eq:nilp}
& \underset{\mathbf{\tau}, \rmQ}{\text{minimize}} 
& & \gM(\tau, \rmQ),\\
& \text{subject to} 
& & \gS(\tau, \rmQ) \leq \text{Size Budget}, \\
&&& \tau \in [0, 1], \\
&&&  \rmQ \in [1, 32]^{C\times B} \cap \sZ^{C\times B}.
\end{aligned}
\end{equation}
Here, $\gM(\cdot)$ and $\gS(\cdot)$ are the quality loss and the estimated size of compressed model under the configuration of $\{\tau, \rmQ\}$, respectively. ``Mixed'' in MINLP means that the variable types include both discrete ($\rmQ$) and continuous ($\tau$) variables. A detailed introduction of MINLP are proposed in the supplementary material.

\vspace{1mm}\noindent\textbf{Structure Decoupling.} Since both $\gM(\cdot)$ and $\gS(\cdot)$ are nonlinear functions of $\{\tau, \rmQ\}$, and the search space is extremely large, solving this problem directly is highly challenging. For example, even when $C=8$ and $B=8$, the discrete combinatorial space already contains $32^{64}$ possibilities—far beyond astronomical scale! Moreover, optimizing the continuous variable $\tau$ within this vast space further increases the difficulty. Besides, existing MINLP solvers~\cite{scip,baronsolver} often require hours to find a reasonable solution. 

However, we observe that only $\tau$ is continuous among these variables, while all others are integers. Besides, by fixing $\tau$, we find that the compressed size varies approximately linearly with the bit-width setting $\rmQ$. Here, the 8-bit value on the horizontal axis indicates that all quantization groups use an 8-bit width. Additionally, we found that the file size under the 16-bit setting is approximately twice that of the 8-bit setting, suggesting that the 16-bit file size can be estimated based on the 8-bit file size. Hence, if we fix the value of $\tau$ and approximate $\gM$ and $\gS$ as functions that are linearizable with respect to $\rmQ$, then this problem can be decoupled into two problems: 1) using discrete sampling to search and fix the value of $\tau$, and then 2) solve the $\rmQ$ via ILP. The advantage of such a decoupling lies in the efficiency of recent ILP solvers~\cite{roy2020pulp,fastdog2022}, which just requires a few seconds.

\begin{algorithm}[t]
    \small
    \SetKwFunction{ILPSearch}{\texttt{01-ILP}}
    \KwIn{Size budget $S_T$ and a pre-trained 3DGS model}
    \KwOut{Hyperparameter set $\Phi^{\ast} = \{\tau^{\ast}, \rmQ^{\ast}\}$}
    Initialize bit-width setting $\rmQ \leftarrow \{8\}^{C \times B}$\;
    Initialize the best quality $M^{\ast} \leftarrow 0$\;
    \For{$\tau \in  \{\tau_1, \tau_2, \tau_3, ...\}$}{
        Compress model with current hyperparameters, obtain size $S_a$\;
        \If{$2 \times S_a < S_T$}{
            {\bf continue}\;
        }
        \While{true}{
            $S_{\Delta} \leftarrow S_a - S_T$\;
            \tcc{\ILPSearch($\cdot$) solver takes $\rmQ$ and $S_{\Delta}$ as input.}
            Search bit-widths via 01-ILP: $\rmQ \leftarrow$ \ILPSearch{$\mathrm{Q}, S_{\Delta}$}\;
            Compress model with $\tau$ and updated $\rmQ$, obtain new size $S_a$\;
            \If{$\frac{|S_a - S_T|}{|S_T|} < 0.05$}{
                {\bf break}\;
            }
        }
        \tcc{Retain the one with the best quality.}
        \If{$\Omega(\rmQ) > M^{\ast}$}{
            $M^{\ast} \leftarrow \Omega(\rmQ)$\;
            $\Phi^{\ast} \leftarrow \{\tau, \rmQ\}$\;
        }
    }
    {\bf return} {$\Phi^{\ast}$}\;
    \caption{\bf Hyperparameter Optimization}
    \label{algo:searching}
\end{algorithm}

\subsection{Solve the MINLP}\label{sec:ilp}

\textbf{Discrete Sampling of $\tau$.} Based on the above observations, we propose to traverse the value of reserve ratio $\tau$. Then, we can search the bit-width settings $\rmQ$ by solving the ILP. Hence, the hyperparameter searching algorithm for size-aware 3D Gaussian compression are proposed in Algo.~\ref{algo:searching}. The input is a target size and a pretrained 3DGS, and the output is the optimal hyperparameters set $\Phi^{\ast} = \{\tau^{\ast}, \rmQ^{\ast}\}$ that can satisfy the size constraint while maximizing visual quality. We iterate over all possible values of $\tau$. Here, $2 \times S_a < S_T$ refers to the situation where, if doubling the bit-width setting under the current configuration (up to a maximum value of 16) cannot achieve the target size, it indicates that $\tau$ is too small, and thus, more Gaussian points need to be retained.
For the bit-width setting search, we formulate the problem as a 0--1 ILP task. The ILP solver takes $\rmQ$ and $S_{\Delta}$ as inputs, where $\rmQ$ serves as the initialization for each search iteration, enabling faster convergence, and $S_{\Delta}$ is used to calibrate the estimated size for more accurate results. After each iteration, the newly determined bit-width setting is used to store the results. If the stored result size $S_a$ is sufficiently close to the target size $S_T$, the search is terminated. For each pair of $\{\tau, \rmQ\}$, we compute the estimated quality $\Omega$ and retain the pair that yields the highest quality. We will introduce $\Omega$ below.

\vspace{1mm}\noindent\textbf{Binary ILP for $\rmQ$.} The input of mixed precision quantization is the important attributes $\gA$, which can be seen as a 2D-matrix. We divide this 2D-matrix into $C \times B$ blocks and there are $Q$ quantization options for each attribute block (e.g., 16 options for 1--16 bits). The search space of the ILP problem is $(C \times B)^Q$. 
The objective of solving the ILP is to find the best bit configuration in this search space that optimally balances quality loss $\Omega$ and the a size limit $\gS$. Besides, we define the bit-width variables as $\rmQ \in \{0, 1\}^{C \times B \times Q}$, which means that we use a one-hot vector $v \in \{0, 1\}^{Q}, |v| = 1$ to represent the bit-width setting for an attribute block. In all, the 0--1 ILP model tries to find the right bit-width setting $\rmQ$ can be formulated as
\begin{equation}
\begin{aligned}
\label{eq:ilp}
    & \underset{\rmQ}{\text{minimize}} 
    & & \Omega(\rmQ),\\
    & \text{subject to} 
    & & \gS(\rmQ) \leq \text{Size Budget}, \\
    &&&  \forall (i, j) \in [0, C) \times [0, B),  \sum_{q=1}^{Q} \rmQ_{i,j,q} = 1.
\end{aligned}
\end{equation}
Here, $\gS(\rmQ)$ and $\Omega(\rmQ)$ denote the estimated file size and estimated quality loss under a bit-width setting $\rmQ$, respectively. Next, we introduce the details of the $\Omega$ and $\gS$.

\vspace{1mm}\noindent\textbf{Hierarchical Solver.} As there are $(C \times B)^Q$ options for $\rmQ$, if we set $(C, B, Q)$ to $(10, 60, 16)$, a typically setting for compressing 3DGS, then there are $2.96 \times 10^{722}$ options. In such a giant search space, figuring out a suitable bit-width setting is time-consuming. Neither of GPU-based~\cite{fastdog2022} or CPU-based~\cite{roy2020pulp} solver can solve the this in minutes. Hence, to accelerate the solving process, we solve this 0--1 ILP problem with two steps. As shown in the left bottom of Fig.~\ref{fig:overview}, at the first step, we search the channel-level bit-width setting $\rmQ_c \in [1, 16]^C$. Then, we calculate the size budget for each channel of attributes based on the $\rmQ_c$. For example, the size budget of channel $i$ can be computed by: $S_c = S_T \frac{\rmQ_{c, i}}{\sum{\rmQ_{c,i}}}$. At the second step, we solve the group-level bit-width setting $\rmQ_g \in [0, 16]^B$ for each channel based on the size limit $S_c$.

\vspace{1mm}\noindent\textbf{Acceleration.} We propose the following techniques to accelerate the problem solving. Since computing the quality metric like PSNR requires traversing the entire training set and evaluating the metric, it takes at least 10 seconds, which significantly slows down the search process. To solve this, we use quantization loss $\Omega(\cdot)$ to replace the metric function $\gM(\cdot)$. 
Here, we assume that the bit-width of each group are independent of one another.
This allows us to precompute the quantization loss of each group of attributes separately, and it only requires $Q$ times quantizations. As for the metric of quality loss, we use the distance between the original attributes and the restored attributes\footnote{Similar assumption can be found in~\cite{dong2019hawq,dong2019hawqv2}.}. Formally, we can precompute the estimated quality loss matrix $\Omega \in \mathbb{R}^{C \times B \times Q}$ with:
\begin{equation}
    \Omega(i, j, b) = | \hat{\mathcal{A}}_{i,j}^b - \mathcal{A}_{i, j} |.
\end{equation}
The $|\cdot|$ can be 1-norm, 2-norm, or $\infty$-norm. We plot the relationship between PNSR and $\Omega$ in the supplementary material, which reveals that minimizing $\Omega$ is equal to maximizing PSNR.
Besides, we set $Q$ as 16 to prune the search space. Finally, we implement a CUDA kernel to accelerate the quantization process, in which each quantization group is quantized in parallel, as shown in Fig.~\ref{fig:overview}.

\label{sec:estimator}
\vspace{1mm}\noindent\textbf{Size Estimator.} A compresssed 3DGS file constains these components: 1) voxelized coordinates, 2) quantized attributes, and 3) metadata, which is used to restore the coordinates and attributes. Fortunately, we can store the voxelized coordinates and metadata to obtain the accurate compressed size in a few seconds. Then the challenge of size estimator lies in estimating the size of quantized attributes. This is to say, we only have to establish a analytical relationship between the compressed file size and the bit-width settings, the actual size of other components can be obtained by saving them to storage. Besides, the size estimator must be a linear function of the bit-width variables.

According to information theory~\cite{Elements_of_information_theory}, the lower bound of bit consumption can be calculated by $ \tau N \times (-\sum_{i}p_i \log_2{p_i})$. However, such a size estimator is not suitable for the formulation of ILP. The reason is that, for each searching iteration, we have to quantize the attributes into integers with the candidate bit-width setting and calculate the probability of each values to derive the bit consumption, which costs a lot of time. Moreover, the relationship between the bit-width settings and the estimated size is non-linear, which cannot satisfy the linear requirement of the ILP. Hence, an explicit and linear relationship between the size $\gS$ and the bit-width setting $\rmQ$ must be established.
Thus, we estimate the size by
\begin{equation}
    \begin{aligned}
        \label{eq:size_estim}
            \gS(\rmQ) = \sum_{i,j} \rmP_{ij} \rmQ_{ij} + \gC + S_\Delta.
    \end{aligned}
    \vspace{-6pt}
\end{equation} 
Here, $\rmP \in \mathbb{R}^{C \times B}$ refers to size of quantization groups. $\mathcal{C}$ refers to the accurate storage consumption of the metadata and the coordinates, which can be obtained by storing them to the disk directly. This process is very fast.
Of course, such a estimation for the compressed file size is not accurate. To calibrate it, we update the $S_\Delta$ multiple times, as shown in Algo.~\ref{algo:searching}.

\begin{figure*}[t]
  \centering
  \includegraphics[width=0.99\linewidth]{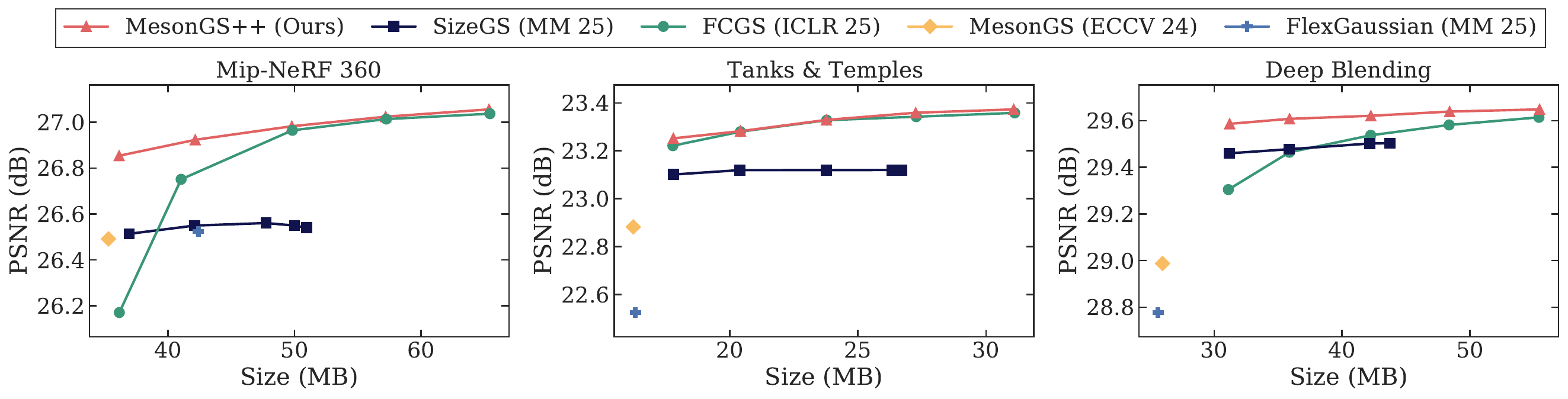}
  \caption{\textbf{Rate-distortion curves for post-training compression.} MesonGS++ achieves better performance across diverse datasets and size budgets.}
  \vspace{-10pt}
  \label{fig:off_comp}
\end{figure*}

\subsection{Entropy Encoding}
The final file contains the following components:
(1) Octree, i.e., quantized 3D coordinates; 
(2) DC coefficients and quantized coefficients; 
(3) Codebook and the corresponding mapping table, or the preserved original $1^{+}$ SH coefficients under big size budget; 
(4) Metadata: Min-Max values of each block of quantized coefficients, octree depth, number of blocks, and bit-width settings.
Notably, the DC coefficients, codebook, and metadata are stored in floating-point format, whereas other components are saved as integers. We compress the octree with G-PCC, and the remained elements are coded and packed via \texttt{torchac}~\cite{mentzer2019practical} and LZ77~\cite{ziv1977universal,ziv1978compression}.

We also provide a fine-tuning module for practical convenience. Specifically, we directly quantize the coordinates and attributes without applying any transformations. To improve compression quality, we fine-tune the model for multiple epochs after each of the point pruning and coordinate quantization steps to restore reconstruction quality. In the final round of fine-tuning, the coordinates are fixed because, when G-PCC is used to compress coordinates, the decompressed coordinates are unordered and do not correspond to the original attribute ordering. To align coordinates and attributes, we compute the Morton order based on the coordinates after the second-stage fine-tuning to jointly sort both, and then construct the quantization groups based on the attributes. Consequently, modifying the coordinate values after quantization may alter the Morton order, disrupting the previously constructed quantization groups and invalidating the searched bit-width settings.
\section{Experiments} 
\label{sec:exp}

\begin{figure*}[t]
    \centering
    \includegraphics[width=0.99\linewidth]{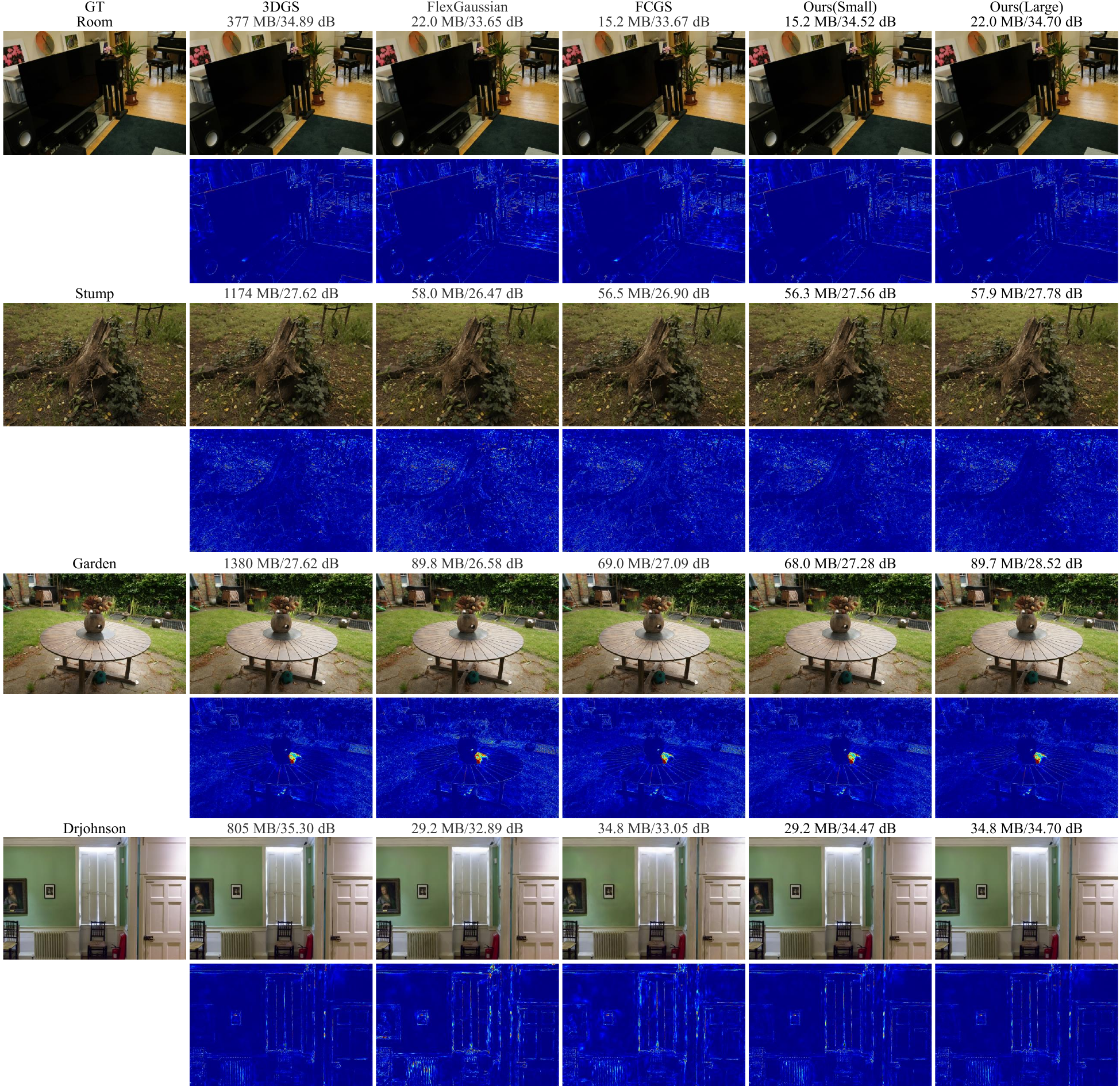}
    \vspace*{-5pt}
    \caption{\textbf{Qualitative comparison.} MesonGS++ preserves visual fidelity after compression, producing renderings that closely match the original 3DGS model. Compared with competing post-training codecs, our method better preserves reflective regions and fine local structures while introducing fewer visible artifacts. Notably, on the \textit{Stump} and \textit{Garden} scenes, our method not only achieves higher PSNR than the original 3DGS but also attains compression rates of 20$\times$ and 15$\times$, respectively.}
    \label{fig:vis_main}
\end{figure*}

We conduct extensive experiments to evaluate rate-distortion performance, target-size control, and visual fidelity under the post-training setting. Compared with existing post-training 3DGS compression methods, MesonGS++ consistently achieves favorable compression-quality trade-offs across diverse size budgets, while preserving rendering quality and satisfying target storage constraints more accurately.

\vspace{1mm}\noindent\textbf{Datasets and Compression Settings.} 
(1) \textbf{Mip-NeRF 360.} The Mip-NeRF 360 dataset~\cite{barron2022mip} contains five outdoor and four indoor scenes. 
Each scene contains 100 to 300 images. 
We use the images at 1600$\times$1063. 
(2) \textbf{Tank\&Temples.} This dataset~\cite{tandt2017} contains two scenes, including \textit{train} and \textit{truck}.
(3) \textbf{Deep Blending.} This dataset~\cite{hedman2018db} contains two scenes, including \textit{drjohnson} and \textit{playroom}.
We evaluate rendering quality and compression performance using PSNR, 
SSIM~\cite{wang2004image}, LPIPS~\cite{zhang2018unreasonable}, and size.
If not otherwise specified, all the metrics are evaluated on the test set.
When calculating the important score, we only use the training dataset.
To obtain the pre-trained 3D Gaussians for compression, we train $30,000$ iterations and 
then save the checkpoints for both datasets. We set the background as white.
Our code is built on the SplatWizard~\cite{liu2025splatwizard}.

\vspace{1mm}\noindent\textbf{Baselines.} 
We compare our method with state-of-the-art post-training compression methods, including MesonGS~\cite{xie2024mesongs}, FCGS~\cite{fcgs2024}, SizeGS~\cite{xie2024sizegs}, and FlexGaussian~\cite{tian2025flexgaussian}.
Although FCGS employs a feed-forward compressor, it operates on the same input as post-training methods, i.e., a pre-trained 3DGS model, making the comparison fair.

\subsection{Main Results}

\vspace{1mm}\noindent\textbf{Rate Distortion Curves.}
As shown in Fig.~\ref{fig:off_comp}, MesonGS++ consistently achieves better compression-quality trade-offs than prior post-training methods across diverse datasets and size budgets.
These results verify that combining an effective post-training codec with size-aware hyperparameter configuration improves both rate-distortion performance and target-size controllability.

\vspace{1mm}\noindent\textbf{Visualization Results.} 
In Fig.~\ref{fig:vis_main}, we present the rendering results and the corresponding error maps. MesonGS++ better preserves reflective regions and fine local structures than competing post-training codecs, while keeping the rendered appearance visually close to the original model. Notably, on the \textit{Stump} and \textit{Garden} scenes, MesonGS++ not only achieves higher PSNR than the original 3DGS but also attains compression rates of 20$\times$ and 15$\times$, respectively.

\begin{figure}[t]
  \centering
   \includegraphics[width=0.99\linewidth]{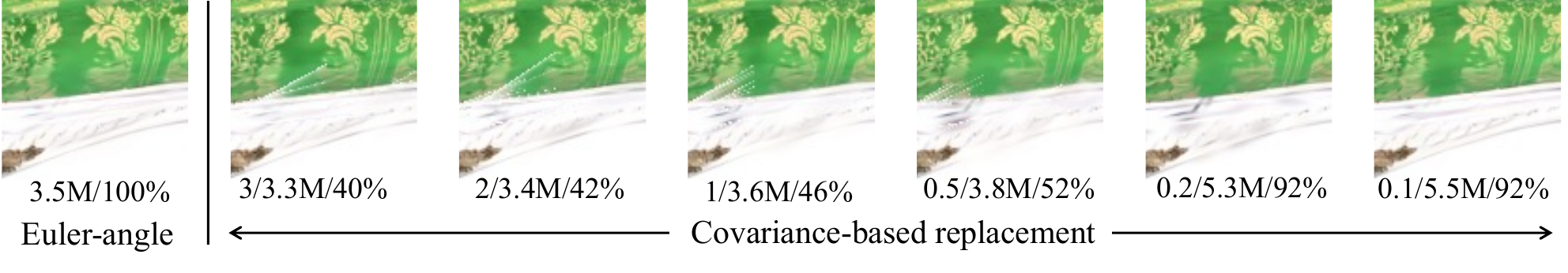}
   \caption{\textbf{Euler-angle-based vs. Covariance-based.} 
   ``A/B/C'' refers to the ``$\lambda_{\mathbf{c}}$ / the size of the compressed file / the percent of positive-definite covariance matrices''. Replacing scales and rotations with covariance leads to white line artifacts, which greatly affects the visual effect.
     We adjust the final file size by compressing a portion of the covariance using the $\lambda_{\mathbf{c}}$.}
   \label{fig:euler_vs_cov}
\end{figure}

\subsection{Ablation Study}

\vspace{1mm}\noindent\textbf{Efficiency and Latency of Different Stages.} We conduct an experiment to demonstrate the benefit of each module in post-training compression.
We calculate the size after a zip compression for fair comparison. 
As shown in Tab.~\ref{tab:diffstage}, compared to the uncompressed baseline, Pruning achieves 
5$\times$ compression but causes a significant PSNR drop for bounded 360 scenes.
However, such a drop is slighter for unbounded scenes.
The following stages are all influenced by the pruning stage.
Of course, all of these stages, instead of Replacement and RAHT, have caused 
varying degrees of damage to the rendering quality.
We can see that the Replacement step is indeed a free lunch of the 3D-GS attribute compression.
We also evaluate the latency of each part of the size-aware hyperparameter search.
Computing the estimated quality is very fast due to our parallel group-wise quantization implementation, only cost 0.49 s. The subsequent calibration (15.10s) and 0-1 ILP stages (58.52) take comparable time, and together (104.51) provide an efficient way to search the bit-width configuration under a prescribed size budget. 

\begin{table}[t]\footnotesize
  \center
  \caption{\textbf{Ablation study of different stages.} ``HS'' refers to searching hyperparameters as MINLP.}
    \label{tab:diffstage}
    \resizebox{\linewidth}{!}{%
  \begin{tabular}{@{}l|ccccc|ccccc@{}}
  \hline
  \multirow{2}{*}{Stages} & \multicolumn{5}{c|}{Synthetic-NeRF} & \multicolumn{5}{c@{}}{Mip-NeRF 360} \\
  \cline{2-11} 
  & \multicolumn{1}{c}{PSNR} & \multicolumn{1}{c}{SSIM} & \multicolumn{1}{c}{LPIPS} & \multicolumn{1}{c}{Size} & \multicolumn{1}{c|}{Latency (s)}
  & \multicolumn{1}{c}{PSNR} & \multicolumn{1}{c}{SSIM} & \multicolumn{1}{c}{LPIPS}& \multicolumn{1}{c}{Size} & \multicolumn{1}{c@{}}{Latency (s)} \\
  \hline
    3D-GS                & 33.37     & 0.9696 & 0.0305 & 68.55     & -- & 28.98     & 0.8647 & 0.1931 & 641.73    & -- \\
+Prune                  & 30.52     & 0.9597 & 0.0383 & 20.99     & 22.73 & 28.69     & 0.8612 & 0.1970 & 265.94    & 165.98 \\
+Voxel                  & 30.44     & 0.9592 & 0.0388 & 20.72     & 5.98 & 28.68     & 0.8610 & 0.1971 & 260.58    & 238.78 \\
+Replace                & 30.44     & 0.9592 & 0.0388 & 20.36     & 0.00 & 28.68     & 0.8610 & 0.1971 & 255.61    & 0.02 \\
+Cluster                & 29.71     & 0.9513 & 0.0469 & 5.63      & 96.75 & 27.74     & 0.8427 & 0.2187 & 79.35     & 267.66 \\
+RAHT                & 29.71     & 0.9513 & 0.0469 & 5.63      & 	0.04 & 27.74     & 0.8427 & 0.2187 & 79.35     & 0.19 \\
+HS                     & 29.47     & 0.9476 & 0.0511 & 1.14      & 111.83 & 27.20     & 0.8238 & 0.2402 & 18.43     & 253.35 \\
  \hline  
\end{tabular}
    }
\end{table}

\vspace{1mm}\noindent\textbf{Composition of Final Storage.}
In the Synthetic-NeRF dataset, the proportions of octree, metadata, important attributes, and unimportant attributes are: 43\%, 0.04\%, 34\%, 23\%, respectively. In the Mip-NeRF 360 dataset, the proportions of the above four elements are: 39\%, 0.02\%, 47\%, 14\%, respectively.

\begin{figure}[t]
  \centering
  \includegraphics[width=\linewidth]{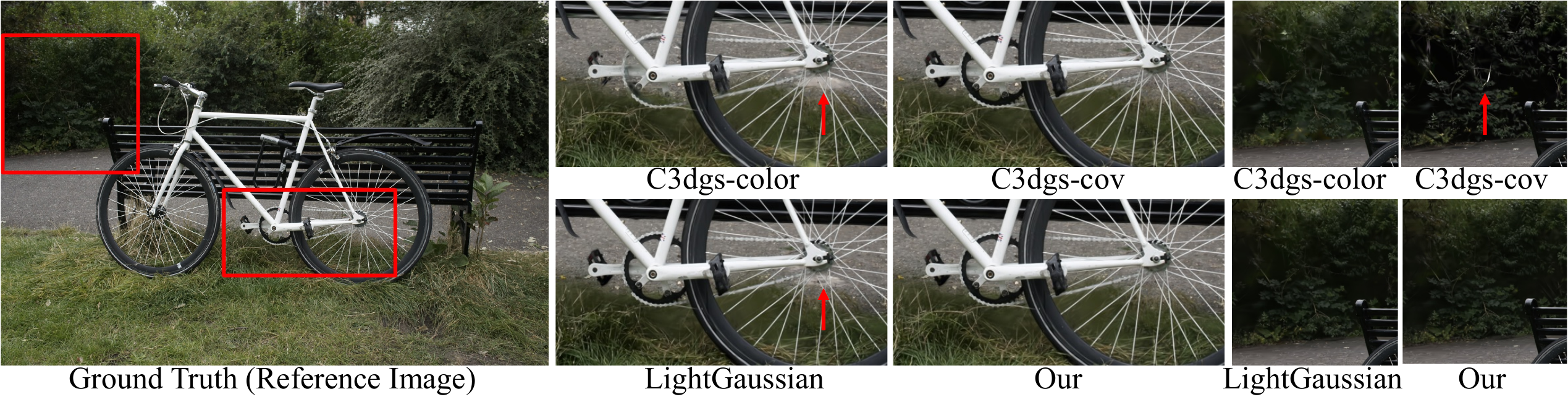}
  \caption{\textbf{Qualitative comparison of different pruning strategies.} Our startegy shows less artifacts.}
  \label{fig:r3q1}
\end{figure}

\vspace{1mm}\noindent\textbf{Pruning Strategy.} Here we compare our pruning strategies with LightGaussian \cite{fan2023lightgaussian} and C3DGS \cite{niedermayr2023compressed}. To achieve fair comparison,
we prune 66\% of the Gaussians for all methods. Tab.~\ref{tab:prune} shows
the superiority of our pruning strategy.
Moreover, Tab.~\ref{tab:prune} indicates that it is necessary to 
incorporate view-dependent importance score $I_d$ in pruning procedure.
Note that the output sizes of these strategies are same, because 
they prune the same percent of Gaussians.
We also provide a qualitative comparison in Fig.~\ref{fig:r3q1}.

\begin{table}[t]\footnotesize
  \centering
  \caption{\textbf{Quantitative comparison on pruning strategy.} Our pruning strategy performs better on average. The best results are in \textbf{bold}, and the second best are \underline{underlined}.}
  \label{tab:prune}
  \resizebox{\columnwidth}{!}{
  \begin{tabular}{@{}l|ccc|ccc|c@{}}
    \hline
    \multirow{2}{*}{Methods}
    & \multicolumn{3}{c|}{Synthetic-NeRF}  & \multicolumn{3}{c|}{Mip-NeRF 360 } & \multirow{2}{*}{Average PSNR} \\
    \cline{2-7}
    & \multicolumn{1}{c}{PSNR} & \multicolumn{1}{c}{SSIM} & \multicolumn{1}{c|}{LPIPS}
    & \multicolumn{1}{c}{PSNR} & \multicolumn{1}{c}{SSIM} & \multicolumn{1}{c|}{LPIPS} \\
    \hline
    C3DGS-color \cite{niedermayr2023compressed} & \textbf{30.79} & \textbf{0.9606} & \textbf{0.0375} & 22.89 & 0.7852 & 0.2471 & 26.84 \\
    C3DGS-cov \cite{niedermayr2023compressed} & 24.37 & 0.9104 & 0.0738 & 15.04 & 0.6992 & 0.3044 & 19.71 \\
    LightGaussian \cite{fan2023lightgaussian} & 26.75 & 0.9372 & 0.0568 & \underline{26.93} & \underline{0.8327} & \underline{0.2290} & 26.84 \\
    \hline
    Our($I_i$) & 30.39 & 0.9591 & 0.0389 & 27.01 &  0.8411 & 0.2164 & \underline{28.70} \\
    Our($I_iI_d$) & \underline{30.52} & \underline{0.9597} & \underline{0.0383} & \textbf{27.52} & \textbf{0.8441} & \textbf{0.2139} & \textbf{29.02} \\
    \hline
  \end{tabular}
  }
\end{table}

\vspace{1mm}\noindent\textbf{Replacement Strategy.} 
We set the bit width as 16 and show the close-up rendering results of Euler-angle based replacement and covariance-based replcaement in the left of Fig.~\ref{fig:euler_vs_cov}. 
We adjust the final file size by compressing a portion of the covariance.
We can see that there are some white line artifacts in the right of covariance-based strategy.
The reason is that the lots of covariance matrices are not 
positive definite after the quantization, i.e., $92\%$ for Euler angle-based vs. $\sim 50\%$ for covariance-based replacement.

\begin{figure}[t]
  \centering
  \includegraphics[width=0.99\linewidth]{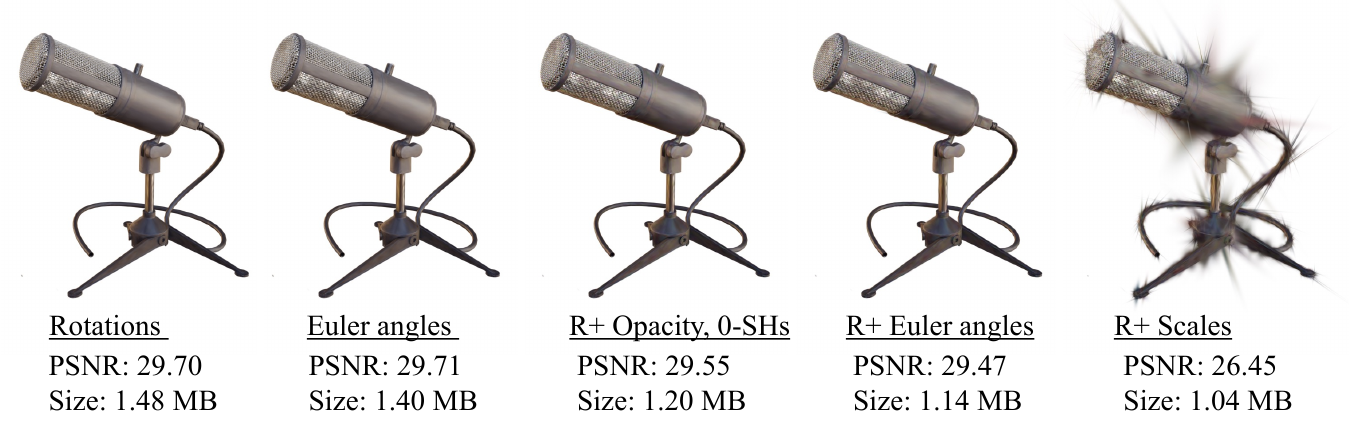}
  \vspace{-10pt}
  \caption{\textbf{Visual results of different attribute transformation stages.} 
  The first image from the left shows the baseline, which means saving the rotations quaternions in the final storage. 
  The second image shows the rendering results after replacing Rotation quaternions with Euler angles.
  ``R+*'' refers to applying RAHT to *. Severe degradation in rendering quality occurrs after applying RAHT to Scales.}
  \label{fig:raht_ab}
\end{figure}

\vspace{1mm}\noindent\textbf{RAHT and Quantization.} For 8-bit quantization, we recommend to not apply RAHT for scales. Due to the activation function of the scale is an exponential function, it is more sensitive than other attributes.
The empirical evidences are shown in Fig.~\ref{fig:raht_ab}.
The information loss caused by the operation of RAHT + Quantization 
is greater than only using Quantization, 
and the exponential function amplifies this error, 
leading to severe performance degradation.

\vspace{1mm}\noindent\textbf{Group-wise Attribute Quantization.} 
As shown in Tab.~\ref{tab:imp}, when employing the per-channel quantization strategy and lowering the pruning threshold from 66\% to 50\%, the compressed file size expands while the rendering quality diminishes. This decline in quality stems from the amplified information loss linked to the extended channel length. As the number of Gaussian points rises, the information loss further intensifies. In contrast, the block quantization method fixes the length of the quantization unit, thereby reducing information loss while providing more flexibility.

\begin{figure}[t]
  \centering
  \includegraphics[width=0.99\linewidth]{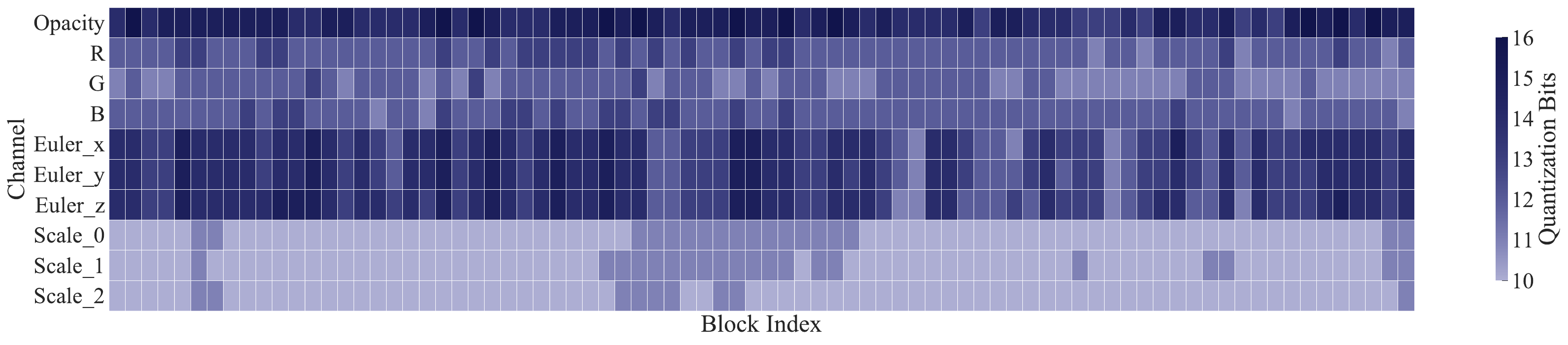}
  \caption{\textbf{Bit widths of quantization groups.} The heatmap visualizes the bit-width assigned to each quantization group. With 10 channels and 80 blocks, there are 800 groups in total. The experiment is conducted on the \textit{bicycle} scene.}
  \label{fig:bit_widths}
\end{figure}

\begin{figure}[t]
  \centering
  \subfloat[{\small Group vs. Channel}\label{fig:diff_comp}]{
      \includegraphics[width=0.47\linewidth]{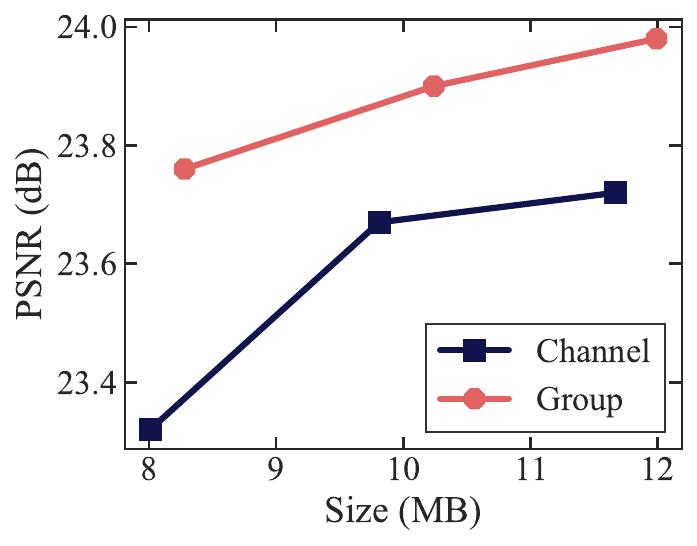}
  }
  \hfill
  \subfloat[{\small Different norms for $\Omega$.}\label{fig:aba_loss}]{
      \includegraphics[width=0.47\linewidth]{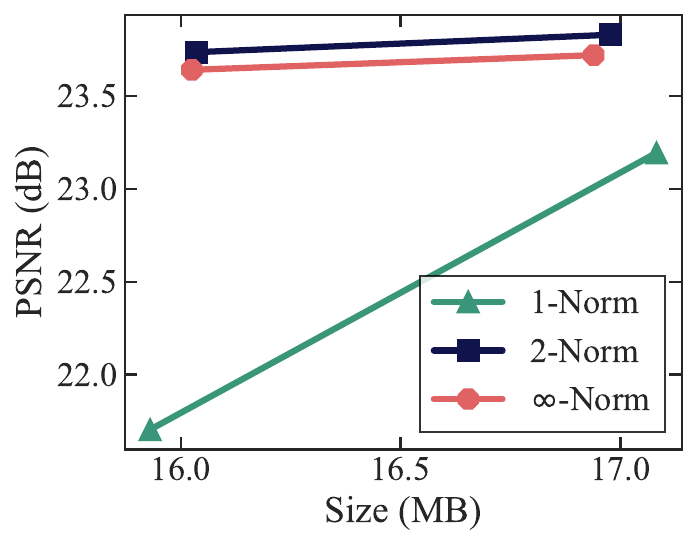}
  }
  \caption{\textbf{Ablation studies.} (a) Results were conducted with sufficient finetuning, confirming that mixed bit-width settings enhance the upper bound of compression quality. (b) 2- and $\infty$-norms significantly outperform 1-norm.}
  \label{fig:ablation}
\end{figure}

\vspace{1mm}\noindent\textbf{Effectiveness of Mixed Bit-width Setting.} An alternative is letting the groups that belong to the same channel share the same bit-width, like the first step of hierarchical solver in Sec.~\ref{sec:ilp}. To verify the necessity of group-wise bit-width setting, we compare the effectiveness of channel-wise and group-wise mixed bit-widths setting in Fig.~\ref{fig:diff_comp}. The results are conducted with enough finetuning. The results show that the performance is improved by fine-grained bit-width setting. In Fig.~\ref{fig:bit_widths}, we visualize the bit-widths of all quantization groups.
We observe that opacity and Euler-angle rotations are typically assigned higher bit-widths, whereas the 0th-order SH coefficients (RGB) and scales tend to use lower bit-widths. This pattern indicates that opacity and rotations are more sensitive to quantization and therefore require higher precision to preserve visual quality.

\begin{table}[t]
\center
\caption{\textbf{The advantage of block quantization.}
    By fixing the length of the vector requiring quantization, block quantization prevents quantization from becoming a performance bottleneck and provides more flexibility.}
  \label{tab:imp}
  \resizebox{\linewidth}{!}{
  \begin{tabular}{@{}l|c|cccc|cccc@{}}
  \hline
\multirow{2}{*}{Strategy} & \multirow{2}{*}{$\tau$} & \multicolumn{4}{c|}{Synthetic-NeRF}      & \multicolumn{4}{c}{Mip-NeRF 360}        \\
\cline{3-10}
                          &                            & PSNR(dB) & SSIM   & LPIPS  & Size(MB) & PSNR(dB) & SSIM   & LPIPS  & Size(MB) \\
\hline
\multirow{2}{*}{Channel}  & 66\%                       & 29.47     & 0.9476 & 0.0511 & 1.14      & 25.30     & 0.7533 & 0.3074 & 11.64     \\
                          & 50\%                       & 30.65     & 0.9529 & 0.0475 & 1.59      & 25.35     & 0.7461 & 0.3147 & 16.47     \\
                          \hline
\multirow{2}{*}{Block}    & 66\%                       &  29.60  &    0.9494    &    0.0490	     &    1.21      & 26.28     & 0.8035 & 0.2598 & 12.46     \\
                          & 50\%                       & 30.97     & 0.9560 & 0.0441 & 1.73      & 27.20     & 0.8238 & 0.2402 & 18.42  \\   
\hline
\end{tabular}
}
\end{table}

\begin{table}[t]
  \centering
  \caption{\textbf{Superiority of 0-1 ILP.} ``GA'': Genetic Algorithm. With up to 16 bit-width choices, the Vanilla ILP and GA that are widely adopted in model quantization methods are unable to quickly search for suitable mixed-precision settings.}
  \resizebox{\linewidth}{!}{
  \begin{tabular}{@{}l|c|ccc@{}}
      \hline
      Method                 & Budget (B)           & Searched (B) & $\Delta$ size (B) & Information loss \\
      \hline
      GA & \multirow{3}{*}{$3\times 10^7$} & 21,833,128        & 8,166,872      & 42,821,038         \\
      Vanilla ILP           &                           & 28,934,805        & 1,065,195      & 1,258,394          \\
      0-1 ILP (Our)               &                           & 29,831,203        & 168,797        & 11,826      \\
      \hline
  \end{tabular}
  }
  \label{tab:ilp_aba}
\end{table}

\vspace{1mm}\noindent\textbf{0-1 ILP Superiority in Searching Bit-widths.} In solving the optimal bit-width setting for different attribute channels, we also demonstrate the superiority of the 0-1 ILP. As shown in Tab.~\ref{tab:ilp_aba}, we experiment with widely-used General ILP~\cite{yao2021hawq} and genetic algorithms~\cite{guo2020single,tang2024retraining}, both of which proved inferior. The 0-1 ILP fully utilizes the size budget while minimizing information loss. General ILP involves variables ranging from 1 to 16. In contrast, 0-1 ILP’s binary values offer finer control, easier integration of constraints, and more efficient solution techniques. Genetic algorithms, though suited for black-box problems, handle constraints less efficiently, making them unsuitable for our problem.

\vspace{1mm}\noindent\textbf{Importance Threshold $\tau$.} 
Pruning is a necessary step for 3D Gaussians compression. 
In Tab.~\ref{tab:imp}, under the bit width of 8, we observe that 
pruning 33\% of points resulted in lower performance compared to pruning 66\% of points.
The reason for this counterintuitive performance drop is that the bit depth of the quantization step is too low, 
and pruning 33\% of points in the 3D-GS model of unbounded scenes still leaves too many points, 
resulting in excessive information loss after quantization. 
Therefore, the quantization bit-width is the bottleneck of overall performance here. 
In Tab.~\ref{tab:imp}, after increasing the quantization bit-width to 16, 
the performance of the 33\% surpasses other baselines.

\vspace{1mm}\noindent\textbf{Estimated Quality Loss $\Omega$.} There are many ways to calculate the distance between the original attributes and the restored attributes. To investigate which metric is better, in Fig.~\ref{fig:aba_loss}, we show the impact of different metrics on the information loss of the final rendering results. We present the PNSR-Size curves under three metrics, including 1-norm, 2-norm, and $\infty$-norm. It can be seen that the performance of 2-norm and $\infty$-norm are nearly the same, both of which outperform 1-norm by a significant margin.

\begin{table}[t]
    \centering
    \caption{\textbf{Robustness evaluation.} The bit-width setting can adapt to different values of the number of blocks $K$, ensuring the visual quality within a given size is not affected by $K$.}
    \label{tab:robust}
    \resizebox{\linewidth}{!}{
    \begin{tabular}{@{}l|c|ccc|c@{}}
    \hline
    $K$  & Budget (MB)  & PSNR  (dB)$\uparrow$  & SSIM  $\uparrow$ &  LPIPS $\downarrow$ & Searched Size (MB) $\downarrow$ \\
    \hline
    40        & \multirow{3}{*}{30} & 25.13  & 0.7410 & 0.2684 & 29.85          \\
    30        &                        & 25.15  & 0.7411 & 0.2685 & 29.91          \\
    50        &                        & 25.14  & 0.7413 & 0.2686 & 29.86          \\
    \hline
    40        & \multirow{3}{*}{20} & 25.07   & 0.7353 & 0.2752  & 19.83          \\
    30        &                        & 25.07   & 0.7357 & 0.2757  & 19.85          \\
    50        &                        & 25.12   & 0.7368 & 0.2743  & 19.92          \\
    \hline
    \end{tabular}
    }
\end{table}

\vspace{1mm}\noindent\textbf{Robustness Evaluation}. We evaluated the size and corresponding performance of the searching algorithm under different numbers of blocks. As shown in Tab.~\ref{tab:robust}, for varying numbers of blocks and different target sizes, our method consistently finds appropriate bit-width settings, ensuring that the final file size is close to the target while maintaining optimal visual quality. Regardless of the block number setting, the final file size and performance are similar, indicating that our method is robust to the number of blocks.

\subsection{Applications and Transferability}

\noindent\textbf{Size-aware Compression on 3DGS Variants.}~We further evaluate the size-aware compression capability of MesonGS++ and the efficiency of its fine-tuning strategy on representative 3DGS variants, including ScaffoldGS~\cite{lu2024scaffold} and 4DGS~\cite{yang2024real}. Table~\ref{tab:dynamic} reports the rendering quality and final storage cost after sufficient fine-tuning. The final model sizes obtained by our search deviate from the prescribed budgets by only 0.8\% for ScaffoldGS and 0.5\% for 4DGS, demonstrating the accuracy of the proposed size estimator. The ScaffoldGS experiments are conducted on the Deep Blending dataset~\cite{hedman2018db}. For 4DGS, MesonGS++ reaches the target budget within 51 s and further improves rendering quality after 28 s of fine-tuning. The 4DGS experiments are conducted on the \textit{flame\_steak} scene from the N3DV dataset~\cite{Li_2022_CVPR}. These results indicate that MesonGS++ is not restricted to the original 3DGS representation; it also transfers effectively to compact and dynamic Gaussian variants.

\begin{table}[t]
  \centering
  \caption{\textbf{Quantitative results on 3DGS variants.}~``CP Time'' refers to post-training compression time. ``FT Time'' refers to finetune time.}
  \label{tab:dynamic}
  \resizebox{\linewidth}{!}{%
  \begin{tabular}{@{}l|c|cccc@{}}
      \hline
      Method    & Budget & PSNR & Size & CP Time (s) & FT Time (s)     \\
      \hline
      ScaffoldGS & - & 29.42 & 663.9 MB & - & - \\
      Ours+ScaffoldGS & 8 MB & 30.24 & 7.92 MB & 447 & 816 \\
      \hline
      4DGS & - & 32.06 & 5.10 GB & - & - \\
      Ours+4DGS & 200 MB & 32.07 & 198.64 MB & 79 & 28 \\
      \hline
  \end{tabular}%
  }
\end{table}

\begin{figure}[t]
  \centering
  \subfloat[\textit{Convergence.}\label{fig:meson_hac_converge}]{%
      \includegraphics[width=0.48\linewidth]{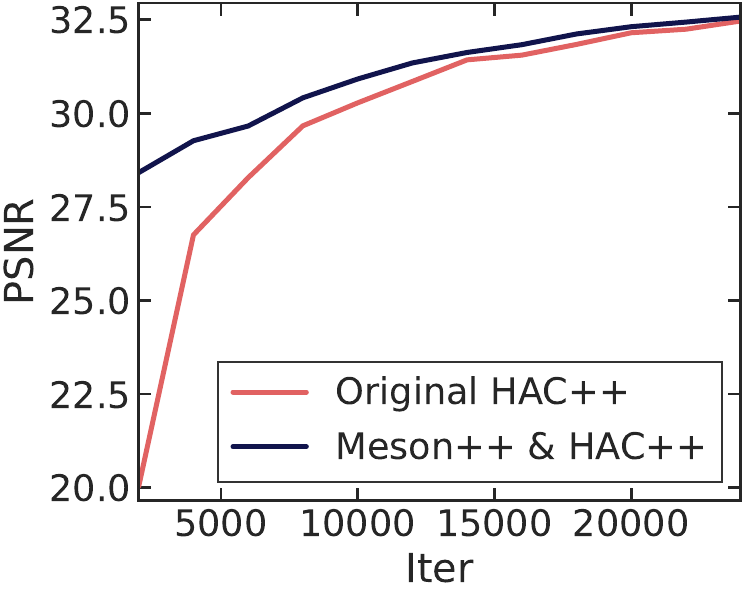}%
  }
  \hfill
  \subfloat[\textit{Ablation.}\label{fig:aba_meson_init}]{%
      \includegraphics[width=0.48\linewidth]{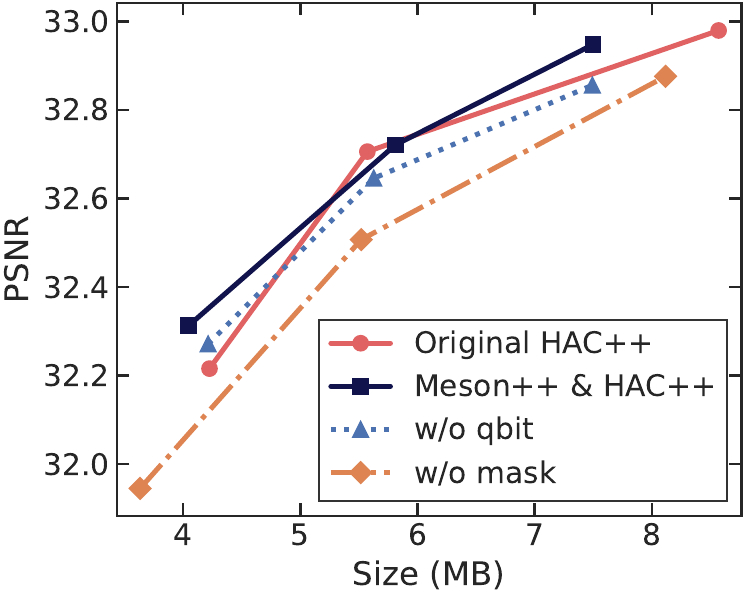}%
  }
  \caption{\textbf{Transferability to online 3DGS compression.} (a) Transferring the compression configuration from MesonGS++ helps HAC++ converge faster without sacrificing final accuracy. (b) Ablation over the transferred priors verifies the effectiveness of \textbf{params}, \textbf{mask}, and \textbf{qbits}, and shows that MesonGS++-based initialization achieves performance comparable to the original HAC++.}
  \label{fig:meson_init}
\end{figure}

\vspace{1mm}\noindent\textbf{Transferability to Online 3DGS Compression.} We investigate transferring three types of priors from MesonGS++ to HAC++, a famous online 3DGS compression method. First, \textbf{params} loads anchor positions, offsets, features, scales, rotations, and opacities to preserve the coarse geometry and latent state. Second, \textbf{mask} initializes pruning logits from MesonGS++ importance scores rather than hard binary decisions: the normalized scores are percentile-clipped to $[\tau_l,\tau_h]$, linearly mapped to $[p_{\min}, p_{\max}]$, and converted to logits via $\ell=\log \frac{p}{1-p}$. Third, \textbf{qbits} converts the searched bit-widths into target quantization steps for the feature, scaling, and offset groups, and uses them to initialize the bias terms of the three $Q$-prediction heads in the grid MLP. Together, \textbf{params}, \textbf{mask}, and \textbf{qbits} provide state, sparsity, and quantization priors, respectively. As shown in Fig.~\ref{fig:meson_hac_converge} and Fig.~\ref{fig:aba_meson_init}, these priors help HAC++ converge faster without sacrificing accuracy, while achieving the same final performance as the HAC++.

\section{Related Work}
\label{sec:rw}
\vspace{1mm}\noindent\textbf{3DGS Compression.} Early compression methods mainly focus on the original 3DGS representation~\cite{niedermayr2023compressed,lee2023compact,girish2023eagles,morgenstern2023compact,navaneet2023compact3d,papantonakis2024i3d,fan2023lightgaussian,xie2024mesongs,wang2024rdo,liu2024compgs,wu2024implicitgs,tang2025neuralgs,xu2025learning}. C3DGS~\cite{niedermayr2023compressed} introduces attribute-sensitivity scoring, replaces quaternion-scale parameterization with covariance, and applies separate vector quantization to geometry and appearance. Lee \textit{et al.}~\cite{lee2023compact} combine learning-based pruning, residual vector quantization for scales and rotations, and NeRF-based coding of SH coefficients. LightGaussian~\cite{fan2023lightgaussian} further integrates octree-based geometry coding, opacity/volume-aware pruning, SH distillation, and vector quantization. Compact3D~\cite{navaneet2023compact3d}, EAGLES~\cite{girish2023eagles}, SOGS~\cite{morgenstern2023compact}, and ReducingGS~\cite{papantonakis2024i3d} respectively explore vector quantization, latent autoencoding, smooth 2D attribute structuring, and mixed-band SH design for additional bitrate reduction. More recent studies have also considered compressing efficient GS variants~\cite{hac2024,lee2025opt,liu2024hemgs} derived from compact representations~\cite{fang2024mini,lu2024scaffold,ali2024trimming,ren2024octree,zhan2025catdgs}. In particular, ScaffoldGS~\cite{lu2024scaffold} organizes anchors into voxels and predicts neural Gaussian attributes from anchor features. Building on this representation, HAC~\cite{hac2024,hac++2025} uses coordinate-conditioned context for quantization and entropy coding, ContextGS~\cite{wang2024contextgs} introduces hierarchical progressive anchor coding, and HEMGS~\cite{liu2024hemgs} incorporates pretrained PointNet++~\cite{qi2017pointnet} to strengthen context modeling. FCGS~\cite{fcgs2024} is a concurrent feed-forward approach that enables one-shot compression, but it is limited to fixed target rates for specific pre-trained models. Orthogonally, layered codec designs~\cite{pcgs2025,disario2025gode,shi2024lapisgs} address rate adaptation under fluctuating network conditions, while recent post-training methods such as POTR~\cite{ramlot2026potr} further broaden the design space for 3D Gaussian compression. Although effective, many existing methods require nontrivial retraining or distillation, which can make them less suitable for resource-constrained scenarios and less compatible with 3DGS extensions~\cite{qin2023langsplat}.

MesonGS++ differs from previous post-training methods by co-designing the codec and its size-aware configuration mechanism. It combines joint importance-based pruning, attribute transformation, and group-wise mixed-precision quantization, while jointly optimizing the reserve ratio and bit-width allocation under a storage budget. Compared with prior post-training codecs~\cite{fan2023lightgaussian,niedermayr2023compressed,tian2025flexgaussian}, our method provides an explicit global control scheme over complicated hyperparameters.

\vspace{1mm}\noindent\textbf{Mixed Precision Quantization.}
Mixed Precision quantization (MPQ) is a widely-used technique to improve the trade-off between the accuracy and efficiency of neural networks~\cite{wang2018haq,dong2019hawq,dong2019hawqv2,yao2021hawq,tang2022mixed}. The challenge with this approach is to find the right mixed-precision setting for the different layers of neural networks. A brute force approach is not feasible since the search space is exponentially large in the number of layers. HAQ~\cite{wang2018haq} employed reinforcement learning to search this space. However, this RL-based solution requires tremendous computational resources. HAWQ~\cite{dong2019hawq,dong2019hawqv2,yao2021hawq} proposed to assign each layer a sensitivity score with the Hessian spectrum and then use an ILP solver to generate mixed-precision settings with various constraints (such as model size and latency). Though CA-NeRF~\cite{liu2024content} uses the MPQ scheme, their method cannot be applied to 3DGS. 

The bit-width setting $\rmQ$ in our framework is similar to the deep learning models. The optimization for the bit-width setting is closely related to mixed precision quantization that mentioned above. Unlike the HAWQ3~\cite{yao2021hawq} that uses the ILP, we use binary ILP to search a better results for the bit-width settings of 3DGS compression. Though some works~\cite{hac2024,liu2024compgs} employed MPQ, their required retraining for configuring the quantization settings.

\vspace{1mm}\noindent\textbf{NeRF Compression.} 
Neural Radiance Field (NeRF) compression primarily targets grid-based NeRF \cite{chen2023factor,chen2023neurbf,barron2023zipnerf,hu2023Tri-MipRF}. The ``grid'' can be voxel-grids~\cite{sun2022direct,DBLP:conf/cvpr/Fridovich-KeilY22,cnc2024}, tri-planes~\cite{DBLP:conf/eccv/ChenXGYS22}, point clouds \cite{xu2022point}, or hash grids \cite{muller2022instant}. Though grid-based NeRF achieves great acceleration, it introduces huge storage requirements, leading to the emergence of NeRF compression works. VQAD~\cite{DBLP:conf/siggraph/TakikawaET0MJF22} proposes to 
generalize different versions of a NeRF with hierarchical coding methods. \cite{deng2023compressing, zhao2023tinynerf, li2023compressing} propose to use frequency domain transformation to reduce the storage demand of the voxel grids.
SHACIRA \cite{girish2023shacira} and CAwa-NeRF \cite{mahmoud2023cawa} compress the hash grid of InstantNGP \cite{muller2022instant}.
BiRF \cite{shin2023binary} introduces a binary quantization scheme.
Masked-wavelet-NeRF \cite{Rho_2023_CVPR} and ACRF \cite{fang2024acrf} 
adopt wavelet transform and RAHT.
ReRF \cite{wang2023neural} is proposed to compress the dynamic NeRF.
To accelerate rendering, some works~\cite{DBLP:journals/tog/ReiserSVSMGBH23, DBLP:conf/iccv/HedmanSMBD21, DBLP:conf/cvpr/ChenFHT23} quantize features to 8 bits and save them as 2D images.

\vspace{1mm}\noindent\textbf{Point Cloud Compression.}
Point cloud compression consists of geometry compression and attribute compression.
The goal of geometry compression is to compress the 3D coordinates of points. Existing methods \cite{meagher1982geometric,schwarz2018emerging}
typically use octree to organize coordinates. 
Attribute compression generally comprises three steps: transform coding, quantization, and entropy coding. 
Transform coding involves designing a careful transformation of attributes 
into the frequency domain to minimize signal redundancy. 
For instance, \cite{zhang2014point} constructs a graph from the point cloud and applies the graph fourier transform to attributes. 
\cite{de2016compression} introduces Haar wavelet transforms to attribute compression. Quantization is used to convert coefficients from transform coding into transmitted symbols
and reduce the high frequency components. 
Entropy coding \cite{huffman1952method,witten1987arithmetic,weinberger2000loco,richardson2004h} aims to encode these symbols into a bitstream.
3DAC \cite{fang20223dac} and Song \textit{et al.} \cite{song2023efficient} propose learning-based entropy models to further reduce the size.

\section{Conclusion}
We present \emph{MesonGS++}, a size-aware post-training codec for 3D Gaussian compression. Beyond effective compression operators, MesonGS++ tackles a key practical bottleneck: globally configuring the many coupled hyperparameters across the pruning--transformation--quantization--encoding pipeline under a prescribed budget. The codec integrates importance-based pruning, octree geometry coding, attribute transformation, selective vector quantization for higher-degree spherical harmonics, and group-wise mixed-precision quantization with entropy coding. A size-aware configuration module jointly optimizes the reserve ratio and bit allocation via discrete sampling, 0-1 ILP, a linear size estimator, and parallel quantization. An optional piecewise finetuning stage further restores quality. Experiments show strong rate-distortion performance, accurate size control, and favorable speed-quality trade-offs, with the searched configurations also serving as useful priors for other compressed Gaussian representations.

\section*{Acknowledgments}
This work is supported in part by National Natural Science Foundation of China (Grant No. 92467204 and 62472249), and Shenzhen Science and Technology Program (Grant No. JCYJ20220818101014030 and KJZD20240903102300001).

\bibliographystyle{IEEEtran}
\bibliography{IEEEabrv,main.bib}

\begin{thebibliography}{100}
\providecommand{\url}[1]{#1}
\csname url@samestyle\endcsname
\providecommand{\newblock}{\relax}
\providecommand{\bibinfo}[2]{#2}
\providecommand{\BIBentrySTDinterwordspacing}{\spaceskip=0pt\relax}
\providecommand{\BIBentryALTinterwordstretchfactor}{4}
\providecommand{\BIBentryALTinterwordspacing}{\spaceskip=\fontdimen2\font plus
\BIBentryALTinterwordstretchfactor\fontdimen3\font minus
  \fontdimen4\font\relax}
\providecommand{\BIBforeignlanguage}[2]{{%
\expandafter\ifx\csname l@#1\endcsname\relax
\typeout{** WARNING: IEEEtran.bst: No hyphenation pattern has been}%
\typeout{** loaded for the language `#1'. Using the pattern for}%
\typeout{** the default language instead.}%
\else
\language=\csname l@#1\endcsname
\fi
#2}}
\providecommand{\BIBdecl}{\relax}
\BIBdecl

\bibitem{luo2025vrdoh}
\BIBentryALTinterwordspacing
Z.~Luo, Z.~Cui, S.~Luo, M.~Chu, and M.~Li, ``Vr-doh: Hands-on 3d modeling in
  virtual reality,'' \emph{ACM Trans. Graph.}, vol.~44, no.~4, Jul. 2025.
  [Online]. Available: \url{https://doi.org/10.1145/3731154}
\BIBentrySTDinterwordspacing

\bibitem{jiang2024vr}
Y.~Jiang, C.~Yu, T.~Xie, X.~Li, Y.~Feng, H.~Wang, M.~Li, H.~Lau, F.~Gao,
  Y.~Yang \emph{et~al.}, ``Vr-gs: A physical dynamics-aware interactive
  gaussian splatting system in virtual reality,'' in \emph{ACM SIGGRAPH 2024
  conference papers}, 2024, pp. 1--1.

\bibitem{yao2025sd}
W.~Yao, S.~Xie, L.~Li, W.~Zhang, Z.~Lai, S.~Dai, K.~Zhang, and Z.~Wang,
  ``Sd-gs: Structured deformable 3d gaussians for efficient dynamic scene
  reconstruction,'' \emph{arXiv preprint arXiv:2507.07465}, 2025.

\bibitem{Zhang_2025_CVPR}
W.~Zhang, S.~Xie, C.~Ren, S.~Xie, C.~Tang, S.~Ge, M.~Wang, and Z.~Wang, ``Evos:
  Efficient implicit neural training via evolutionary selector,'' in
  \emph{Proceedings of the IEEE/CVF Conference on Computer Vision and Pattern
  Recognition (CVPR)}, June 2025, pp. 30\,472--30\,482.

\bibitem{sympowerinr}
W.~Zhang, S.~Xie, C.~Ren, S.~Ge, M.~Wang, and Z.~Wang, ``Enhancing implicit
  neural representations via symmetric power transformation,'' in
  \emph{{AAAI}}.\hskip 1em plus 0.5em minus 0.4em\relax {AAAI} Press, 2025, pp.
  10\,157--10\,165.

\bibitem{zhang2025understanding}
W.~Zhang, B.~Li, S.~Xie, C.~Ren, Y.~Xue, and Z.~Wang, ``Understanding bias
  terms in neural representations,'' in \emph{Advances in Neural Information
  Processing Systems (NeurIPS)}, 2025.

\bibitem{gu2024dragscene}
C.~Gu, Z.~Li, Z.~Zhang, Y.~Bai, S.~Xie, and Z.~Wang, ``Dragscene: Interactive
  3d scene editing with single-view drag instructions,'' \emph{arXiv preprint
  arXiv:2412.13552}, 2024.

\bibitem{zhou2024drivinggaussian}
X.~Zhou, Z.~Lin, X.~Shan, Y.~Wang, D.~Sun, and M.-H. Yang, ``Drivinggaussian:
  Composite gaussian splatting for surrounding dynamic autonomous driving
  scenes,'' in \emph{Proceedings of the IEEE/CVF conference on computer vision
  and pattern recognition}, 2024, pp. 21\,634--21\,643.

\bibitem{yan2024street}
Y.~Yan, H.~Lin, C.~Zhou, W.~Wang, H.~Sun, K.~Zhan, X.~Lang, X.~Zhou, and
  S.~Peng, ``Street gaussians: Modeling dynamic urban scenes with gaussian
  splatting,'' in \emph{European Conference on Computer Vision}.\hskip 1em plus
  0.5em minus 0.4em\relax Springer, 2024, pp. 156--173.

\bibitem{Wang_2025_CVPR}
Y.~Wang, P.~Yang, Z.~Xu, J.~Sun, Z.~Zhang, Y.~Chen, H.~Bao, S.~Peng, and
  X.~Zhou, ``Freetimegs: Free gaussian primitives at anytime anywhere for
  dynamic scene reconstruction,'' in \emph{Proceedings of the IEEE/CVF
  Conference on Computer Vision and Pattern Recognition (CVPR)}, June 2025, pp.
  21\,750--21\,760.

\bibitem{zhao2025real2edit2real}
Y.~Zhao, H.~Fan, D.~Chen, S.~Chen, L.~Chen, X.~Li, G.~Ren, and H.~Dong,
  ``Real2edit2real: Generating robotic demonstrations via a 3d control
  interface,'' \emph{arXiv preprint arXiv:2512.19402}, 2025.

\bibitem{fan2025twinaligner}
H.~Fan, H.~Dai, J.~Zhang, J.~Li, Q.~Yan, Y.~Zhao, M.~Gao, J.~Wu, H.~Tang, and
  H.~Dong, ``Twinaligner: Visual-dynamic alignment empowers physics-aware
  real2sim2real for robotic manipulation,'' \emph{arXiv preprint
  arXiv:2512.19390}, 2025.

\bibitem{gu2025igen}
C.~Gu, H.~Kang, J.~Lin, J.~Wang, D.~Wu, S.~Xie, F.~Huang, J.~Ge, Z.~Gong, L.~Li
  \emph{et~al.}, ``Igen: Scalable data generation for robot learning from
  open-world images,'' \emph{arXiv preprint arXiv:2512.01773}, 2025.

\bibitem{kerbl20233d}
B.~Kerbl, G.~Kopanas, T.~Leimk{\"u}hler, and G.~Drettakis, ``3d gaussian
  splatting for real-time radiance field rendering,'' \emph{ACM Transactions on
  Graphics (ToG)}, vol.~42, no.~4, pp. 1--14, 2023.

\bibitem{barron2022mip}
J.~T. Barron, B.~Mildenhall, D.~Verbin, P.~P. Srinivasan, and P.~Hedman,
  ``Mip-nerf 360: Unbounded anti-aliased neural radiance fields,'' in
  \emph{Proceedings of the IEEE/CVF Conference on Computer Vision and Pattern
  Recognition}, 2022, pp. 5470--5479.

\bibitem{niedermayr2023compressed}
S.~Niedermayr, J.~Stumpfegger, and R.~Westermann, ``Compressed 3d gaussian
  splatting for accelerated novel view synthesis,'' in \emph{Proceedings of the
  IEEE/CVF Conference on Computer Vision and Pattern Recognition (CVPR)}, 2024.

\bibitem{fan2023lightgaussian}
Z.~Fan, K.~Wang, K.~Wen, Z.~Zhu, D.~Xu, and Z.~Wang, ``{LightGaussian:
  Unbounded 3D Gaussian Compression with 15x Reduction and 200+ FPS},'' in
  \emph{Advances in neural information processing systems (NeurIPS)}, 2024.

\bibitem{navaneet2023compact3d}
K.~Navaneet, K.~P. Meibodi, S.~A. Koohpayegani, and H.~Pirsiavash, ``Compgs:
  Smaller and faster gaussian splatting with vector quantization,''
  \emph{ECCV}, 2024.

\bibitem{morgenstern2023compact}
W.~Morgenstern, F.~Barthel, A.~Hilsmann, and P.~Eisert, ``Compact 3d scene
  representation via self-organizing gaussian grids,'' in \emph{European
  Conference on Computer Vision}.\hskip 1em plus 0.5em minus 0.4em\relax
  Springer, 2024.

\bibitem{jc2023gsplat}
J.~Červený, \url{https://gsplat.tech}, 2023, accessed: 2023-10-28.

\bibitem{ap2023gssmall}
A.~Pranckevičius, ``Making gaussian splats smaller,''
  \url{https://aras-p.info/blog/2023/09/13/Making-Gaussian-Splats-smaller/},
  2023, accessed: 2023-10-28.

\bibitem{ap2023gssmaller}
------, ``Making gaussian splats more smaller,''
  \url{https://aras-p.info/blog/2023/09/27/Making-Gaussian-Splats-more-smaller/},
  2023, accessed: 2023-10-28.

\bibitem{pcgs2025}
Y.~Chen, M.~Li, Q.~Wu, W.~Lin, M.~Harandi, and J.~Cai, ``Pcgs: Progressive
  compression of 3d gaussian splatting,'' \emph{arXiv preprint
  arXiv:2503.08511}, 2025.

\bibitem{zhan2025cat}
Y.-T. Zhan, C.-Y. Ho, H.~Yang, Y.-H. Chen, J.~C. Chiang, Y.-L. Liu, and W.-H.
  Peng, ``Cat-3dgs: A context-adaptive triplane approach to
  rate-distortion-optimized 3dgs compression,'' \emph{arXiv preprint
  arXiv:2503.00357}, 2025.

\bibitem{xie2024mesongs}
S.~Xie, W.~Zhang, C.~Tang, Y.~Bai, R.~Lu, S.~Ge, and Z.~Wang, ``Mesongs:
  Post-training compression of 3d gaussians via efficient attribute
  transformation,'' in \emph{European Conference on Computer Vision}.\hskip 1em
  plus 0.5em minus 0.4em\relax Springer, 2024.

\bibitem{xie2024sizegs}
S.~Xie, J.~Liu, W.~Zhang, S.~Ge, S.~Pan, C.~Tang, Y.~Bai, C.~Zhang, X.~Fan, and
  Z.~Wang, ``Sizegs: Size-aware compression of 3d gaussian splatting via mixed
  integer programming,'' in \emph{ACM MM}, 2025.

\bibitem{ramlot2026potr}
\BIBentryALTinterwordspacing
B.~Ramlot, M.~Courteaux, P.~Lambert, and G.~V. Wallendael, ``Potr:
  Post-training 3dgs compression,'' 2026. [Online]. Available:
  \url{https://arxiv.org/abs/2601.14821}
\BIBentrySTDinterwordspacing

\bibitem{liu2025splatwizard}
\BIBentryALTinterwordspacing
X.~Liu, Y.~Zhou, J.~Wang, Y.~Huang, S.~Xie, S.~Qin, M.~Hong, J.~Li, Y.~Wang,
  Z.~Wang, S.-T. Xia, and B.~Chen, ``Splatwizard: A benchmark toolkit for 3d
  gaussian splatting compression,'' 2025. [Online]. Available:
  \url{https://arxiv.org/abs/2512.24742}
\BIBentrySTDinterwordspacing

\bibitem{zhang2025gaussianspa}
Y.~Zhang, W.~Jia, W.~Niu, and M.~Yin, ``Gaussianspa: An"
  optimizing-sparsifying" simplification framework for compact and high-quality
  3d gaussian splatting,'' in \emph{Proceedings of the Computer Vision and
  Pattern Recognition Conference}, 2025, pp. 26\,673--26\,682.

\bibitem{wu2024implicitgs}
\BIBentryALTinterwordspacing
M.~Wu and T.~Tuytelaars, ``Implicit gaussian splatting with efficient
  multi-level tri-plane representation,'' 2024. [Online]. Available:
  \url{https://arxiv.org/abs/2408.10041}
\BIBentrySTDinterwordspacing

\bibitem{liu2024compgs}
X.~Liu, X.~Wu, P.~Zhang, S.~Wang, Z.~Li, and S.~Kwong, ``Compgs: Efficient 3d
  scene representation via compressed gaussian splatting,'' in
  \emph{Proceedings of the 32nd ACM International Conference on Multimedia},
  2024.

\bibitem{hac++2025}
Y.~Chen, Q.~Wu, W.~Lin, M.~Harandi, and J.~Cai, ``Hac++: Towards 100x
  compression of 3d gaussian splatting,'' \emph{arXiv preprint
  arXiv:2501.12255}, 2025.

\bibitem{hac2024}
------, ``Hac: Hash-grid assisted context for 3d gaussian splatting
  compression,'' in \emph{European Conference on Computer Vision}, 2024.

\bibitem{girish2023eagles}
S.~Girish, K.~Gupta, and A.~Shrivastava, ``Eagles: Efficient accelerated 3d
  gaussians with lightweight encodings,'' in \emph{European Conference on
  Computer Vision}, 2024.

\bibitem{wang2024end}
H.~Wang, H.~Zhu, T.~He, R.~Feng, J.~Deng, J.~Bian, and Z.~Chen, ``End-to-end
  rate-distortion optimized 3d gaussian representation,'' in \emph{European
  Conference on Computer Vision}.\hskip 1em plus 0.5em minus 0.4em\relax
  Springer, 2024, pp. 76--92.

\bibitem{wang2024contextgs}
Y.~Wang, Z.~Li, L.~Guo, W.~Yang, A.~C. Kot, and B.~Wen, ``Contextgs: Compact 3d
  gaussian splatting with anchor level context model,'' in \emph{Advances in
  neural information processing systems (NeurIPS)}, 2024.

\bibitem{papantonakis2024i3d}
\BIBentryALTinterwordspacing
P.~Papantonakis, G.~Kopanas, B.~Kerbl, A.~Lanvin, and G.~Drettakis, ``Reducing
  the memory footprint of 3d gaussian splatting,'' \emph{Proceedings of the ACM
  on Computer Graphics and Interactive Techniques}, vol.~7, no.~1, May 2024.
  [Online]. Available: \url{https://repo-sam.inria.fr/fungraph/reduced-3dgs/}
\BIBentrySTDinterwordspacing

\bibitem{shin2025locality}
S.~Shin, J.~Park, and S.~Cho, ``Locality-aware gaussian compression for fast
  and high-quality rendering,'' \emph{arXiv preprint arXiv:2501.05757}, 2025.

\bibitem{cao2024lightweight}
J.~Cao, V.~Goel, C.~Wang, A.~Kag, J.~Hu, S.~Korolev, C.~Jiang, S.~Tulyakov, and
  J.~Ren, ``Lightweight predictive 3d gaussian splats,'' \emph{arXiv preprint
  arXiv:2406.19434}, 2024.

\bibitem{zhang2024lp}
Z.~Zhang, T.~Song, Y.~Lee, L.~Yang, C.~Peng, R.~Chellappa, and D.~Fan,
  ``Lp-3dgs: Learning to prune 3d gaussian splatting,'' \emph{Advances in
  Neural Information Processing Systems}, vol.~37, pp. 122\,434--122\,457,
  2024.

\bibitem{hanson2025pup}
A.~Hanson, A.~Tu, V.~Singla, M.~Jayawardhana, M.~Zwicker, and T.~Goldstein,
  ``Pup 3d-gs: Principled uncertainty pruning for 3d gaussian splatting,'' in
  \emph{Proceedings of the Computer Vision and Pattern Recognition Conference},
  2025, pp. 5949--5958.

\bibitem{hanson2025speedy}
A.~Hanson, A.~Tu, G.~Lin, V.~Singla, M.~Zwicker, and T.~Goldstein,
  ``Speedy-splat: Fast 3d gaussian splatting with sparse pixels and sparse
  primitives,'' in \emph{Proceedings of the Computer Vision and Pattern
  Recognition Conference}, 2025, pp. 21\,537--21\,546.

\bibitem{niemeyer2025radsplat}
M.~Niemeyer, F.~Manhardt, M.-J. Rakotosaona, M.~Oechsle, D.~Duckworth,
  R.~Gosula, K.~Tateno, J.~Bates, D.~Kaeser, and F.~Tombari, ``Radsplat:
  Radiance field-informed gaussian splatting for robust real-time rendering
  with 900+ fps,'' in \emph{2025 International Conference on 3D Vision
  (3DV)}.\hskip 1em plus 0.5em minus 0.4em\relax IEEE, 2025, pp. 134--144.

\bibitem{fang2024mini}
G.~Fang and B.~Wang, ``Mini-splatting: Representing scenes with a constrained
  number of gaussians,'' in \emph{European Conference on Computer Vision},
  2024.

\bibitem{fang2024mini2}
------, ``Mini-splatting2: Building 360 scenes within minutes via aggressive
  gaussian densification,'' \emph{arXiv preprint arXiv:2411.12788}, 2024.

\bibitem{chen2025megs}
J.~Chen, Y.~Chen, Y.~Zou, Y.~Huang, P.~Wang, Y.~Liu, Y.~Sun, and W.~Wang,
  ``{MEGS}$^2$: Memory-efficient gaussian splatting via spherical gaussians and
  unified pruning,'' \emph{arXiv preprint arXiv:2509.07021}, 2025.

\bibitem{lee2026safeguardgs}
Y.~Lee, Z.~Zhang, and D.~Fan, ``Safeguardgs: 3d gaussian primitive pruning
  while avoiding catastrophic scene destruction,'' in \emph{Proceedings of the
  IEEE/CVF Winter Conference on Applications of Computer Vision}, 2026, pp.
  8479--8489.

\bibitem{huang2025entropygs}
Y.~Huang, J.~Pang, F.~Zhu, and D.~Tian, ``Entropygs: An efficient entropy
  coding on 3d gaussian splatting,'' \emph{arXiv preprint arXiv:2508.10227},
  2025.

\bibitem{liu2025efficientgs}
W.~Liu, T.~Guan, B.~Zhu, L.~Xu, Z.~Song, D.~Li, Y.~Wang, and W.~Yang,
  ``Efficientgs: Streamlining gaussian splatting for large-scale
  high-resolution scene representation,'' \emph{IEEE MultiMedia}, 2025.

\bibitem{tian2025flexgaussian}
\BIBentryALTinterwordspacing
B.~Tian, Q.~Gao, S.~Xianyu, X.~Cui, and M.~Zhang, ``Flexgaussian: Flexible and
  cost-effective training-free compression for 3d gaussian splatting,'' in
  \emph{Proceedings of the 33rd ACM International Conference on Multimedia},
  ser. MM '25.\hskip 1em plus 0.5em minus 0.4em\relax New York, NY, USA:
  Association for Computing Machinery, 2025, p. 7287–7296. [Online].
  Available: \url{https://doi.org/10.1145/3746027.3754744}
\BIBentrySTDinterwordspacing

\bibitem{fcgs2024}
Y.~Chen, Q.~Wu, M.~Li, W.~Lin, M.~Harandi, and J.~Cai, ``Fast feedforward 3d
  gaussian splatting compression,'' \emph{arXiv preprint arXiv:2410.08017},
  2024.

\bibitem{de2016compression}
R.~L. De~Queiroz and P.~A. Chou, ``Compression of 3d point clouds using a
  region-adaptive hierarchical transform,'' \emph{IEEE Transactions on Image
  Processing}, vol.~25, no.~8, pp. 3947--3956, 2016.

\bibitem{equitz1989new}
W.~H. Equitz, ``A new vector quantization clustering algorithm,'' \emph{IEEE
  transactions on acoustics, speech, and signal processing}, vol.~37, no.~10,
  pp. 1568--1575, 1989.

\bibitem{tmc13}
{MPEG Group}, ``{MPEG G-PCC Test Model TMC13},''
  \url{https://github.com/MPEGGroup/mpeg-pcc-tmc13}, 2024, accessed: 2024.

\bibitem{mentzer2019practical}
F.~Mentzer, E.~Agustsson, M.~Tschannen, R.~Timofte, and L.~Van~Gool,
  ``Practical full resolution learned lossless image compression,'' in
  \emph{Proceedings of the IEEE Conference on Computer Vision and Pattern
  Recognition (CVPR)}, 2019.

\bibitem{ewa_volume_splatting}
M.~Zwicker, H.~Pfister, J.~van Baar, and M.~Gross, ``Ewa volume splatting,'' in
  \emph{Proceedings Visualization, 2001. VIS '01.}, 2001, pp. 29--538.

\bibitem{yifan2019differentiable}
W.~Yifan, F.~Serena, S.~Wu, C.~{\"O}ztireli, and O.~Sorkine-Hornung,
  ``Differentiable surface splatting for point-based geometry processing,''
  \emph{ACM Transactions on Graphics (TOG)}, vol.~38, no.~6, pp. 1--14, 2019.

\bibitem{zwicker2001surface}
M.~Zwicker, H.~Pfister, J.~Van~Baar, and M.~Gross, ``Surface splatting,'' in
  \emph{Proceedings of the 28th annual conference on Computer graphics and
  interactive techniques}, 2001, pp. 371--378.

\bibitem{li2023compressing}
L.~Li, Z.~Shen, Z.~Wang, L.~Shen, and L.~Bo, ``Compressing volumetric radiance
  fields to 1 mb,'' in \emph{Proceedings of the IEEE/CVF Conference on Computer
  Vision and Pattern Recognition}, 2023, pp. 4222--4231.

\bibitem{web_scale_cluster}
\BIBentryALTinterwordspacing
D.~Sculley, ``Web-scale k-means clustering,'' in \emph{Proceedings of the 19th
  International Conference on World Wide Web}, ser. WWW '10.\hskip 1em plus
  0.5em minus 0.4em\relax New York, NY, USA: Association for Computing
  Machinery, 2010, p. 1177–1178. [Online]. Available:
  \url{https://doi.org/10.1145/1772690.1772862}
\BIBentrySTDinterwordspacing

\bibitem{frantar-gptq}
E.~Frantar, S.~Ashkboos, T.~Hoefler, and D.~Alistarh, ``{GPTQ}: Accurate
  post-training compression for generative pretrained transformers,''
  \emph{ICLR}, 2023.

\bibitem{2023-emnlp-osplus}
\BIBentryALTinterwordspacing
X.~Wei, Y.~Zhang, Y.~Li, X.~Zhang, R.~Gong, J.~Guo, and X.~Liu, ``Outlier
  suppression+: Accurate quantization of large language models by equivalent
  and effective shifting and scaling,'' in \emph{Proceedings of the 2023
  Conference on Empirical Methods in Natural Language Processing}, H.~Bouamor,
  J.~Pino, and K.~Bali, Eds.\hskip 1em plus 0.5em minus 0.4em\relax Singapore:
  Association for Computational Linguistics, December 2023, pp. 1648--1665.
  [Online]. Available: \url{https://aclanthology.org/2023.emnlp-main.102}
\BIBentrySTDinterwordspacing

\bibitem{MLSYS2024_5edb57c0}
\BIBentryALTinterwordspacing
Y.~Zhao, C.-Y. Lin, K.~Zhu, Z.~Ye, L.~Chen, S.~Zheng, L.~Ceze,
  A.~Krishnamurthy, T.~Chen, and B.~Kasikci, ``Atom: Low-bit quantization for
  efficient and accurate llm serving,'' in \emph{Proceedings of Machine
  Learning and Systems}, P.~Gibbons, G.~Pekhimenko, and C.~D. Sa, Eds., vol.~6,
  2024, pp. 196--209. [Online]. Available:
  \url{https://proceedings.mlsys.org/paper_files/paper/2024/file/5edb57c05c81d04beb716ef1d542fe9e-Paper-Conference.pdf}
\BIBentrySTDinterwordspacing

\bibitem{tang2022mixed}
C.~Tang, K.~Ouyang, Z.~Wang, Y.~Zhu, Y.~Wang, W.~Ji, and W.~Zhu,
  ``Mixed-precision neural network quantization via learned layer-wise
  importance,'' in \emph{European Conference on Computer Vision}, 2022.

\bibitem{scip}
\BIBentryALTinterwordspacing
(2018) {The SCIP Optimization Suite 6.0}. [Online]. Available:
  \url{https://www.scipopt.org/}
\BIBentrySTDinterwordspacing

\bibitem{baronsolver}
\BIBentryALTinterwordspacing
(2025) {BARON Solver}. [Online]. Available:
  \url{https://minlp.com/baron-solver}
\BIBentrySTDinterwordspacing

\bibitem{roy2020pulp}
J.~S. Roy and S.~A. Mitchell, ``{PuLP}: An lp modeler written in python,''
  \url{https://github.com/coin-or/pulp}, 2020, accessed: 2026-04-29.

\bibitem{fastdog2022}
\BIBentryALTinterwordspacing
A.~Abbas and P.~Swoboda, ``Fastdog: Fast discrete optimization on {GPU},'' in
  \emph{{IEEE/CVF} Conference on Computer Vision and Pattern Recognition,
  {CVPR} 2022, New Orleans, LA, USA, June 18-24, 2022}.\hskip 1em plus 0.5em
  minus 0.4em\relax {IEEE}, 2022, pp. 439--449. [Online]. Available:
  \url{https://doi.org/10.1109/CVPR52688.2022.00053}
\BIBentrySTDinterwordspacing

\bibitem{dong2019hawq}
Z.~Dong, Z.~Yao, A.~Gholami, M.~W. Mahoney, and K.~Keutzer, ``{HAWQ}: {H}essian
  {AW}are {Q}uantization of neural networks with mixed-precision,'' in
  \emph{The IEEE International Conference on Computer Vision (ICCV)}, October
  2019.

\bibitem{dong2019hawqv2}
Z.~Dong, Z.~Yao, D.~Arfeen, A.~Gholami, M.~W. Mahoney, and K.~Keutzer,
  ``{HAWQ-V2}: {H}essian aware trace-weighted quantization of neural
  networks,'' in \emph{Advances in neural information processing systems
  (NeurIPS)}, 2020.

\bibitem{Elements_of_information_theory}
T.~M. Cover, \emph{Elements of information theory}.\hskip 1em plus 0.5em minus
  0.4em\relax John Wiley \& Sons, 1999.

\bibitem{ziv1977universal}
J.~Ziv and A.~Lempel, ``A universal algorithm for sequential data
  compression,'' \emph{IEEE Transactions on information theory}, vol.~23,
  no.~3, pp. 337--343, 1977.

\bibitem{ziv1978compression}
------, ``Compression of individual sequences via variable-rate coding,''
  \emph{IEEE transactions on Information Theory}, vol.~24, no.~5, pp. 530--536,
  1978.

\bibitem{tandt2017}
A.~Knapitsch, J.~Park, Q.-Y. Zhou, and V.~Koltun, ``Tanks and temples:
  Benchmarking large-scale scene reconstruction,'' \emph{ACM Transactions on
  Graphics (ToG)}, vol.~36, no.~4, pp. 1--13, 2017.

\bibitem{hedman2018db}
P.~Hedman, J.~Philip, T.~Price, J.-M. Frahm, G.~Drettakis, and G.~Brostow,
  ``Deep blending for free-viewpoint image-based rendering,'' \emph{ACM
  Transactions on Graphics (ToG)}, vol.~37, no.~6, pp. 1--15, 2018.

\bibitem{wang2004image}
Z.~Wang, A.~C. Bovik, H.~R. Sheikh, and E.~P. Simoncelli, ``Image quality
  assessment: from error visibility to structural similarity,'' \emph{IEEE
  transactions on image processing}, vol.~13, no.~4, pp. 600--612, 2004.

\bibitem{zhang2018unreasonable}
R.~Zhang, P.~Isola, A.~A. Efros, E.~Shechtman, and O.~Wang, ``The unreasonable
  effectiveness of deep features as a perceptual metric,'' in \emph{Proceedings
  of the IEEE conference on computer vision and pattern recognition}, 2018, pp.
  586--595.

\bibitem{yao2021hawq}
Z.~Yao, Z.~Dong, Z.~Zheng, A.~Gholami, J.~Yu, E.~Tan, L.~Wang, Q.~Huang,
  Y.~Wang, M.~Mahoney \emph{et~al.}, ``Hawq-v3: Dyadic neural network
  quantization,'' in \emph{International Conference on Machine Learning}.\hskip
  1em plus 0.5em minus 0.4em\relax PMLR, 2021, pp. 11\,875--11\,886.

\bibitem{guo2020single}
Z.~Guo, X.~Zhang, H.~Mu, W.~Heng, Z.~Liu, Y.~Wei, and J.~Sun, ``Single path
  one-shot neural architecture search with uniform sampling,'' in
  \emph{Computer Vision--ECCV 2020: 16th European Conference, Glasgow, UK,
  August 23--28, 2020, Proceedings, Part XVI 16}.\hskip 1em plus 0.5em minus
  0.4em\relax Springer, 2020, pp. 544--560.

\bibitem{tang2024retraining}
C.~Tang, Y.~Meng, J.~Jiang, S.~Xie, R.~Lu, X.~Ma, Z.~Wang, and W.~Zhu,
  ``Retraining-free model quantization via one-shot weight-coupling learning,''
  in \emph{Proceedings of the IEEE/CVF Conference on Computer Vision and
  Pattern Recognition}, 2024, pp. 15\,855--15\,865.

\bibitem{lu2024scaffold}
T.~Lu, M.~Yu, L.~Xu, Y.~Xiangli, L.~Wang, D.~Lin, and B.~Dai, ``Scaffold-gs:
  Structured 3d gaussians for view-adaptive rendering,'' in \emph{Proceedings
  of the IEEE/CVF Conference on Computer Vision and Pattern Recognition}, 2024,
  pp. 20\,654--20\,664.

\bibitem{yang2024real}
Z.~Yang, H.~Yang, Z.~Pan, X.~Zhu, and L.~Zhang, ``Real-time photorealistic
  dynamic scene representation and rendering with 4d gaussian splatting,'' in
  \emph{ICLR}, 2024.

\bibitem{Li_2022_CVPR}
T.~Li, M.~Slavcheva, M.~Zollh\"ofer, S.~Green, C.~Lassner, C.~Kim, T.~Schmidt,
  S.~Lovegrove, M.~Goesele, R.~Newcombe, and Z.~Lv, ``Neural 3d video synthesis
  from multi-view video,'' in \emph{Proceedings of the IEEE/CVF Conference on
  Computer Vision and Pattern Recognition (CVPR)}, June 2022, pp. 5521--5531.

\bibitem{lee2023compact}
J.~C. Lee, D.~Rho, X.~Sun, J.~H. Ko, and E.~Park, ``Compact 3d gaussian
  representation for radiance field,'' \emph{CVPR}, 2024.

\bibitem{wang2024rdo}
H.~Wang, H.~Zhu, T.~He, R.~Feng, J.~Deng, J.~Bian, and Z.~Chen, ``End-to-end
  rate-distortion optimized 3d gaussian representation,'' in \emph{European
  Conference on Computer Vision}, 2024.

\bibitem{tang2025neuralgs}
\BIBentryALTinterwordspacing
Z.~Tang, C.~Feng, X.~Cheng, W.~Yu, J.~Zhang, Y.~Liu, X.~Long, W.~Wang, and
  L.~Yuan, ``Neuralgs: Bridging neural fields and 3d gaussian splatting for
  compact 3d representations,'' 2025. [Online]. Available:
  \url{https://arxiv.org/abs/2503.23162}
\BIBentrySTDinterwordspacing

\bibitem{xu2025learning}
H.~Xu, X.~Wu, and X.~Zhang, ``Learning hierarchical sparse transform coding of
  3dgs,'' \emph{arXiv preprint arXiv:2505.22908}, 2025.

\bibitem{lee2025opt}
\BIBentryALTinterwordspacing
J.~C. Lee, J.~H. Ko, and E.~Park, ``Optimized minimal 3d gaussian splatting,''
  2025. [Online]. Available: \url{https://arxiv.org/abs/2503.16924}
\BIBentrySTDinterwordspacing

\bibitem{liu2024hemgs}
\BIBentryALTinterwordspacing
L.~Liu, Z.~Chen, and D.~Xu, ``{HEMGS:} {A} hybrid entropy model for 3d gaussian
  splatting data compression,'' \emph{CoRR}, vol. abs/2411.18473, 2024.
  [Online]. Available: \url{https://doi.org/10.48550/arXiv.2411.18473}
\BIBentrySTDinterwordspacing

\bibitem{ali2024trimming}
M.~S. Ali, M.~Qamar, S.-H. Bae, and E.~Tartaglione, ``Trimming the fat:
  Efficient compression of 3d gaussian splats through pruning,'' in
  \emph{BMVC}, 2024.

\bibitem{ren2024octree}
K.~Ren, L.~Jiang, T.~Lu, M.~Yu, L.~Xu, Z.~Ni, and B.~Dai, ``Octree-gs: Towards
  consistent real-time rendering with lod-structured 3d gaussians,''
  \emph{arXiv preprint arXiv:2403.17898}, 2024.

\bibitem{zhan2025catdgs}
\BIBentryALTinterwordspacing
Y.-T. Zhan, C.-Y. Ho, H.~Yang, Y.-H. Chen, J.~C. Chiang, Y.-L. Liu, and W.-H.
  Peng, ``{CAT}-3{DGS}: A context-adaptive triplane approach to
  rate-distortion-optimized 3{DGS} compression,'' in \emph{The Thirteenth
  International Conference on Learning Representations}, 2025. [Online].
  Available: \url{https://openreview.net/forum?id=m3KuuE2ozw}
\BIBentrySTDinterwordspacing

\bibitem{qi2017pointnet}
\BIBentryALTinterwordspacing
C.~R. Qi, L.~Yi, H.~Su, and L.~J. Guibas, ``Pointnet++: Deep hierarchical
  feature learning on point sets in a metric space,'' in \emph{Advances in
  Neural Information Processing Systems 30: Annual Conference on Neural
  Information Processing Systems 2017, December 4-9, 2017, Long Beach, CA,
  {USA}}, I.~Guyon, U.~von Luxburg, S.~Bengio, H.~M. Wallach, R.~Fergus,
  S.~V.~N. Vishwanathan, and R.~Garnett, Eds., 2017, pp. 5099--5108. [Online].
  Available:
  \url{https://proceedings.neurips.cc/paper/2017/hash/d8bf84be3800d12f74d8b05e9b89836f-Abstract.html}
\BIBentrySTDinterwordspacing

\bibitem{disario2025gode}
\BIBentryALTinterwordspacing
F.~D. Sario, R.~Renzulli, M.~Grangetto, A.~Sugimoto, and E.~Tartaglione,
  ``Gode: Gaussians on demand for progressive level of detail and scalable
  compression,'' 2025. [Online]. Available:
  \url{https://arxiv.org/abs/2501.13558}
\BIBentrySTDinterwordspacing

\bibitem{shi2024lapisgs}
Y.~Shi, S.~Gasparini, G.~Morin, and W.~T. Ooi, ``{LapisGS}: Layered progressive
  {3D Gaussian} splatting for adaptive streaming,'' in \emph{International
  Conference on 3D Vision, 3DV 2025, Singapore, March 25-28, 2025}.\hskip 1em
  plus 0.5em minus 0.4em\relax {IEEE}, 2025.

\bibitem{qin2023langsplat}
M.~Qin, W.~Li, J.~Zhou, H.~Wang, and H.~Pfister, ``Langsplat: 3d language
  gaussian splatting,'' \emph{arXiv preprint arXiv:2312.16084}, 2023.

\bibitem{wang2018haq}
K.~Wang, Z.~Liu, Y.~Lin, J.~Lin, and S.~Han, ``{HAQ}: Hardware-aware automated
  quantization,'' in \emph{Proceedings of the IEEE/CVF Conference on Computer
  Vision and Pattern Recognition}, 2019.

\bibitem{liu2024content}
W.~Liu, X.~X. Zheng, J.~Yu, and X.~Lou, ``Content-aware radiance fields:
  Aligning model complexity with scene intricacy through learned bitwidth
  quantization,'' in \emph{European Conference on Computer Vision}.\hskip 1em
  plus 0.5em minus 0.4em\relax Springer, 2024, pp. 239--256.

\bibitem{chen2023factor}
\BIBentryALTinterwordspacing
A.~Chen, Z.~Xu, X.~Wei, S.~Tang, H.~Su, and A.~Geiger, ``Factor fields: A
  unified framework for neural fields and beyond,'' 2023. [Online]. Available:
  \url{https://arxiv.org/abs/2302.01226}
\BIBentrySTDinterwordspacing

\bibitem{chen2023neurbf}
Z.~Chen, Z.~Li, L.~Song, L.~Chen, J.~Yu, J.~Yuan, and Y.~Xu, ``Neurbf: A neural
  fields representation with adaptive radial basis functions,'' in
  \emph{Proceedings of the IEEE/CVF International Conference on Computer
  Vision}, 2023, pp. 4182--4194.

\bibitem{barron2023zipnerf}
J.~T. Barron, B.~Mildenhall, D.~Verbin, P.~P. Srinivasan, and P.~Hedman,
  ``Zip-nerf: Anti-aliased grid-based neural radiance fields,'' \emph{ICCV},
  2023.

\bibitem{hu2023Tri-MipRF}
W.~Hu, Y.~Wang, L.~Ma, B.~Yang, L.~Gao, X.~Liu, and Y.~Ma, ``Tri-miprf: Tri-mip
  representation for efficient anti-aliasing neural radiance fields,'' in
  \emph{ICCV}, 2023.

\bibitem{sun2022direct}
C.~Sun, M.~Sun, and H.-T. Chen, ``Direct voxel grid optimization: Super-fast
  convergence for radiance fields reconstruction,'' in \emph{Proceedings of the
  IEEE/CVF Conference on Computer Vision and Pattern Recognition}, 2022, pp.
  5459--5469.

\bibitem{DBLP:conf/cvpr/Fridovich-KeilY22}
\BIBentryALTinterwordspacing
S.~Fridovich{-}Keil, A.~Yu, M.~Tancik, Q.~Chen, B.~Recht, and A.~Kanazawa,
  ``Plenoxels: Radiance fields without neural networks,'' in \emph{{IEEE/CVF}
  Conference on Computer Vision and Pattern Recognition, {CVPR} 2022, New
  Orleans, LA, USA, June 18-24, 2022}.\hskip 1em plus 0.5em minus 0.4em\relax
  {IEEE}, 2022, pp. 5491--5500. [Online]. Available:
  \url{https://doi.org/10.1109/CVPR52688.2022.00542}
\BIBentrySTDinterwordspacing

\bibitem{cnc2024}
Y.~Chen, Q.~Wu, M.~Harandi, and J.~Cai, ``How far can we compress
  instant-ngp-based nerf?'' in \emph{Proceedings of the IEEE/CVF Conference on
  Computer Vision and Pattern Recognition}, 2024.

\bibitem{DBLP:conf/eccv/ChenXGYS22}
\BIBentryALTinterwordspacing
A.~Chen, Z.~Xu, A.~Geiger, J.~Yu, and H.~Su, ``Tensorf: Tensorial radiance
  fields,'' in \emph{Computer Vision - {ECCV} 2022 - 17th European Conference,
  Tel Aviv, Israel, October 23-27, 2022, Proceedings, Part {XXXII}}, ser.
  Lecture Notes in Computer Science, S.~Avidan, G.~J. Brostow, M.~Ciss{\'{e}},
  G.~M. Farinella, and T.~Hassner, Eds., vol. 13692.\hskip 1em plus 0.5em minus
  0.4em\relax Springer, 2022, pp. 333--350. [Online]. Available:
  \url{https://doi.org/10.1007/978-3-031-19824-3\_20}
\BIBentrySTDinterwordspacing

\bibitem{xu2022point}
Q.~Xu, Z.~Xu, J.~Philip, S.~Bi, Z.~Shu, K.~Sunkavalli, and U.~Neumann,
  ``Point-nerf: Point-based neural radiance fields,'' in \emph{Proceedings of
  the IEEE/CVF Conference on Computer Vision and Pattern Recognition}, 2022,
  pp. 5438--5448.

\bibitem{muller2022instant}
T.~M{\"u}ller, A.~Evans, C.~Schied, and A.~Keller, ``Instant neural graphics
  primitives with a multiresolution hash encoding,'' \emph{ACM Transactions on
  Graphics (ToG)}, vol.~41, no.~4, pp. 1--15, 2022.

\bibitem{DBLP:conf/siggraph/TakikawaET0MJF22}
\BIBentryALTinterwordspacing
T.~Takikawa, A.~Evans, J.~Tremblay, T.~M{\"{u}}ller, M.~McGuire, A.~Jacobson,
  and S.~Fidler, ``Variable bitrate neural fields,'' in \emph{{SIGGRAPH} '22:
  Special Interest Group on Computer Graphics and Interactive Techniques
  Conference, Vancouver, BC, Canada, August 7 - 11, 2022}, M.~Nandigjav, N.~J.
  Mitra, and A.~Hertzmann, Eds.\hskip 1em plus 0.5em minus 0.4em\relax {ACM},
  2022, pp. 41:1--41:9. [Online]. Available:
  \url{https://doi.org/10.1145/3528233.3530727}
\BIBentrySTDinterwordspacing

\bibitem{deng2023compressing}
C.~L. Deng and E.~Tartaglione, ``Compressing explicit voxel grid
  representations: fast nerfs become also small,'' in \emph{Proceedings of the
  IEEE/CVF Winter Conference on Applications of Computer Vision}, 2023, pp.
  1236--1245.

\bibitem{zhao2023tinynerf}
T.~Zhao, J.~Chen, C.~Leng, and J.~Cheng, ``Tinynerf: Towards 100 x compression
  of voxel radiance fields,'' in \emph{Proceedings of the AAAI Conference on
  Artificial Intelligence}, 2023, pp. 3588--3596.

\bibitem{girish2023shacira}
S.~Girish, A.~Shrivastava, and K.~Gupta, ``Shacira: Scalable hash-grid
  compression for implicit neural representations,'' in \emph{Proceedings of
  the IEEE/CVF International Conference on Computer Vision}, 2023, pp.
  17\,513--17\,524.

\bibitem{mahmoud2023cawa}
O.~Mahmoud, T.~Ladune, and M.~Gendrin, ``Cawa-nerf: Instant learning of
  compression-aware nerf features,'' \emph{arXiv preprint arXiv:2310.14695},
  2023.

\bibitem{shin2023binary}
S.~Shin and J.~Park, ``Binary radiance fields,'' \emph{arXiv preprint
  arXiv:2306.07581}, 2023.

\bibitem{Rho_2023_CVPR}
D.~Rho, B.~Lee, S.~Nam, J.~C. Lee, J.~H. Ko, and E.~Park, ``Masked wavelet
  representation for compact neural radiance fields,'' in \emph{Proceedings of
  the IEEE/CVF Conference on Computer Vision and Pattern Recognition (CVPR)},
  June 2023, pp. 20\,680--20\,690.

\bibitem{fang2024acrf}
G.~Fang, Q.~Hu, L.~Wang, and Y.~Guo, ``{ACRF}: {C}ompressing explicit neural
  radiance fields via attribute compression,'' in \emph{International
  Conference on Learning Representations(ICLR)}, 2024.

\bibitem{wang2023neural}
L.~Wang, Q.~Hu, Q.~He, Z.~Wang, J.~Yu, T.~Tuytelaars, L.~Xu, and M.~Wu,
  ``Neural residual radiance fields for streamably free-viewpoint videos,'' in
  \emph{Proceedings of the IEEE/CVF Conference on Computer Vision and Pattern
  Recognition}, 2023, pp. 76--87.

\bibitem{DBLP:journals/tog/ReiserSVSMGBH23}
\BIBentryALTinterwordspacing
C.~Reiser, R.~Szeliski, D.~Verbin, P.~P. Srinivasan, B.~Mildenhall, A.~Geiger,
  J.~T. Barron, and P.~Hedman, ``{MERF:} memory-efficient radiance fields for
  real-time view synthesis in unbounded scenes,'' \emph{{ACM} Trans. Graph.},
  vol.~42, no.~4, pp. 89:1--89:12, 2023. [Online]. Available:
  \url{https://doi.org/10.1145/3592426}
\BIBentrySTDinterwordspacing

\bibitem{DBLP:conf/iccv/HedmanSMBD21}
\BIBentryALTinterwordspacing
P.~Hedman, P.~P. Srinivasan, B.~Mildenhall, J.~T. Barron, and P.~E. Debevec,
  ``Baking neural radiance fields for real-time view synthesis,'' in \emph{2021
  {IEEE/CVF} International Conference on Computer Vision, {ICCV} 2021,
  Montreal, QC, Canada, October 10-17, 2021}.\hskip 1em plus 0.5em minus
  0.4em\relax {IEEE}, 2021, pp. 5855--5864. [Online]. Available:
  \url{https://doi.org/10.1109/ICCV48922.2021.00582}
\BIBentrySTDinterwordspacing

\bibitem{DBLP:conf/cvpr/ChenFHT23}
\BIBentryALTinterwordspacing
Z.~Chen, T.~A. Funkhouser, P.~Hedman, and A.~Tagliasacchi, ``Mobilenerf:
  Exploiting the polygon rasterization pipeline for efficient neural field
  rendering on mobile architectures,'' in \emph{{IEEE/CVF} Conference on
  Computer Vision and Pattern Recognition, {CVPR} 2023, Vancouver, BC, Canada,
  June 17-24, 2023}.\hskip 1em plus 0.5em minus 0.4em\relax {IEEE}, 2023, pp.
  16\,569--16\,578. [Online]. Available:
  \url{https://doi.org/10.1109/CVPR52729.2023.01590}
\BIBentrySTDinterwordspacing

\bibitem{meagher1982geometric}
D.~Meagher, ``Geometric modeling using octree encoding,'' \emph{Computer
  graphics and image processing}, vol.~19, no.~2, pp. 129--147, 1982.

\bibitem{schwarz2018emerging}
S.~Schwarz, M.~Preda, V.~Baroncini, M.~Budagavi, P.~Cesar, P.~A. Chou, R.~A.
  Cohen, M.~Krivoku{\'c}a, S.~Lasserre, Z.~Li \emph{et~al.}, ``Emerging mpeg
  standards for point cloud compression,'' \emph{IEEE Journal on Emerging and
  Selected Topics in Circuits and Systems}, vol.~9, no.~1, pp. 133--148, 2018.

\bibitem{zhang2014point}
C.~Zhang, D.~Florencio, and C.~Loop, ``Point cloud attribute compression with
  graph transform,'' in \emph{2014 IEEE International Conference on Image
  Processing (ICIP)}.\hskip 1em plus 0.5em minus 0.4em\relax IEEE, 2014, pp.
  2066--2070.

\bibitem{huffman1952method}
D.~A. Huffman, ``A method for the construction of minimum-redundancy codes,''
  \emph{Proceedings of the IRE}, vol.~40, no.~9, pp. 1098--1101, 1952.

\bibitem{witten1987arithmetic}
I.~H. Witten, R.~M. Neal, and J.~G. Cleary, ``Arithmetic coding for data
  compression,'' \emph{Communications of the ACM}, vol.~30, no.~6, pp.
  520--540, 1987.

\bibitem{weinberger2000loco}
M.~J. Weinberger, G.~Seroussi, and G.~Sapiro, ``The loco-i lossless image
  compression algorithm: Principles and standardization into jpeg-ls,''
  \emph{IEEE Transactions on Image processing}, vol.~9, no.~8, pp. 1309--1324,
  2000.

\bibitem{richardson2004h}
I.~E. Richardson, \emph{H. 264 and MPEG-4 video compression: video coding for
  next-generation multimedia}.\hskip 1em plus 0.5em minus 0.4em\relax John
  Wiley \& Sons, 2004.

\bibitem{fang20223dac}
G.~Fang, Q.~Hu, H.~Wang, Y.~Xu, and Y.~Guo, ``3dac: Learning attribute
  compression for point clouds,'' in \emph{Proceedings of the IEEE/CVF
  Conference on Computer Vision and Pattern Recognition}, 2022, pp.
  14\,819--14\,828.

\bibitem{song2023efficient}
R.~Song, C.~Fu, S.~Liu, and G.~Li, ``Efficient hierarchical entropy model for
  learned point cloud compression,'' in \emph{Proceedings of the IEEE/CVF
  Conference on Computer Vision and Pattern Recognition}, 2023, pp.
  14\,368--14\,377.

\end{thebibliography}
\vspace{20pt}
\noindent\textbf{Shuzhao Xie} is currently a PhD student at Shenzhen International Graduate School, Tsinghua University.
He received his B.S. degree from Beijing Normal University, China, in 2020,
and his M.Eng. degree from Tsinghua University, China, in 2023.
His research focuses on graphics and deep learning.

\vspace{20pt}
\noindent\textbf{Junchen Ge} is currently a master's student at Shenzhen International Graduate School, Tsinghua University.
He received his B.S. degree from Dalian University of Technology, China, in 2025.
His research interests include 3D computer vision and graphics, with applications in gaming and embodied intelligence. His recent work spans human pose estimation, 3D Gaussian splatting, and robust SLAM for robotics.

\vspace{20pt}
\noindent\textbf{Weixiang Zhang} is currently a PhD student at The Hong Kong University of Science and Technology. He received his B.S. degree from Beihang University, China, in 2020, and his M.Eng. degree from Tsinghua University, China, in 2025. His research focuses on efficient deep learning.

\vspace{20pt}
\noindent\textbf{Jiahang Liu} is currently an undergraduate student at Harbin Institute of Technology, Shenzhen, China. His research interests include embodied intelligence and 3D vision.

\vspace{20pt}
\noindent\textbf{Chen Tang} is currently a Ph.D. student at the Multimedia Laboratory (MMLab), The Chinese University of Hong Kong. He received his Master's degree in Computer Technology from Tsinghua University. Before CUHK, he worked as a research assistant in the Department of Computer Science and Technology at Tsinghua University. He was a research intern in the System and Networking Research Group of Microsoft Research Asia (MSRA). His research interests are in efficient deep learning (e.g., quantization, sparsity), hardware-aware neural architecture search, and dynamic neural networks.

\vspace{20pt}
\noindent\textbf{Yunpeng Bai} received his Bachelor's degree in Computer Science and Technology from the Dalian University of Technology, China, in 2020, followed by a Master's degree from Tsinghua University, China, in 2023. He is currently a CS PhD student at UT Austin advised by Qixing Huang.
His research interests include computer vision and computer graphics.

\vspace{20pt}
\noindent\textbf{Shijia Ge} is currently a master’s student at the Shenzhen International Graduate School, Tsinghua University. He received his B.S. degree from Huazhong University of Science and Technology, China, in 2023. His research focuses on 3D vision and robotics.

\vspace{20pt}
\noindent\textbf{Jingyan Jiang} received her Ph.D. degree in Computer Science from Jilin University, China, in 2020. From 2020 to 2022, she was a Postdoctoral Research Fellow at the Shenzhen International Graduate School, Tsinghua University. She is currently an assistant professor with the School of Artificial Intelligence, Shenzhen Technology University, Shenzhen, China. Her research interests mainly focus on model domain generalization and distributed machine learning.

\vspace{20pt}
\noindent\textbf{Yuzhi Huang} received the B.E. degree from Fujian Normal University, China, in 2023. He is currently pursuing the M.S. degree with the School of Informatics, Xiamen University, China. His research interests include vision-language models (VLMs), foundation models, and computer vision.

\vspace{20pt}
\noindent\textbf{Fengnian Yang} is currently a master’s student at Shenzhen International Graduate School, Tsinghua University. He received his B.S. degree from Xidian University, China, in 2024. His research interests lie in 3D vision and embodied intelligence.

\vspace{20pt}
\noindent\textbf{Cong Zhang} received M.Sc. from Zhengzhou University, Zhengzhou, China, in 2012, and Ph.D. from Simon Fraser University, Canada, in 2018. He is currently a visiting researcher at Simon Fraser University. His research interests include deep learning for smart energy, multimedia communications, and cloud/edge computing. His co-authored papers received the Distinguished Paper Award of IEEE ICDCS (2024) and the Best Paper Award Honorable Mention of IEEE Transactions on Multimedia (2016).

\vspace{20pt}
\noindent\textbf{Xiaoyi Fan} received the B.Eng. degree from the
Beijing University of Posts and Telecommunications
in 2013, and the PhD degree from Simon Fraser
University in 2018.
He is currently CTO with Shenzhen Jiangxing Intelligence Inc.
He is a recipient of NSERC Postdoctoral Fellowship
(2019). He is a recipient of the IEEE ICDCS Distinguished Paper Award (2024) and the IWQoS Best
Poster Award (2024). His research interests include
physical AI, edge computing, and smart grid.

\vspace{20pt}
\noindent\textbf{Zhi Wang} is currently an associate professor at Shenzhen International Graduate School, Tsinghua University. He received his Ph.D. in 2014 and his B.E. in 2008, both from Department of Computer Science and Technology, Tsinghua University. His research areas include multimedia networks, mobile cloud computing, and large-scale machine learning systems. He is a recipient of the Natural Science Award of the Ministry of Education (First Prize) in 2017, the National Natural Science Award (Second Prize) in 2018, and the Technology Invention Award of the Chinese Institute of Electronics (First Prize) in 2020. In addition, his research won the Best Paper Award of ACM Multimedia, Best Paper Award of IEEE TMM, the Best Student Paper Award of MMM, and the Best Paper Award of ACM Multimedia, HUMA Workshop. He is an Associate Editor of IEEE TMM and Guest Editor of ACM TIST and JCST.

\newpage
\appendix
\section{More Experiment Results}

We provide MesonGS++'s per-scene results on the Mip-NeRF 360 dataset~\cite{barron2022mip}, the Tank\&Temples~\cite{tandt2017}, and the Deep Blending dataset~\cite{hedman2018db}. Tables~\ref{tab:supp_mip360}, \ref{tab:supp_deep_blending}, and \ref{tab:supp_tandt} report the compressed model size and rendering quality under five target-size profiles, from Large to Tiny. These results show the detailed per-scene trade-off between storage cost and visual quality, where higher PSNR/SSIM and lower LPIPS indicate better reconstruction quality.
These results corresponds to the data points in the RD curves of Fig.~6 in the main paper.

\begin{table}[t]
    \centering
    \caption{\textbf{Detailed results on Mip-NeRF 360.} We report the size and rendering quality of MesonGS++ under five target-size profiles.}
    \label{tab:supp_mip360}
    \begingroup
    \renewcommand{\arraystretch}{0.71}
    \resizebox{\linewidth}{!}{%
    \begin{tabular}{cccccc}
    \hline
    Type                        & Scene    & Size (MB) & PSNR ($\uparrow$) & SSIM ($\uparrow$) & LPIPS ($\downarrow$) \\
    \hline
    \multirow{10}{*}{Large}     & bicycle  & 109.0751  & 24.4798           & 0.735407            & 0.289537             \\
                                & bonsai   & 24.293    & 31.9207           & 0.941863            & 0.243593             \\
                                & counter  & 24.2901   & 28.904            & 0.909689            & 0.252471             \\
                                & flowers  & 72.8299   & 21.232            & 0.580182            & 0.398802             \\
                                & garden   & 112.996   & 26.211            & 0.830662            & 0.180954             \\
                                & kitchen  & 39.1796   & 30.7541           & 0.924736            & 0.152784             \\
                                & room     & 26.5124   & 31.5163           & 0.923454            & 0.275306             \\
                                & stump    & 101.5627  & 26.3179           & 0.757056            & 0.295586             \\
                                & treehill & 77.536    & 22.1635           & 0.624007            & 0.405221             \\
                                & \textbf{AVG} & 65.3639 & 27.0555           & 0.803006            & 0.277140             \\
                                \hline
    \multirow{10}{*}{Medium}    & bicycle  & 95.6088   & 24.4703           & 0.734414            & 0.290847             \\
                                & bonsai   & 21.298    & 31.8358           & 0.94098             & 0.24479              \\
                                & counter  & 21.2983   & 28.8568           & 0.908494            & 0.254131             \\
                                & flowers  & 63.9522   & 21.2201           & 0.579133            & 0.399991             \\
                                & garden   & 98.8334   & 26.1958           & 0.829802            & 0.182632             \\
                                & kitchen  & 34.3881   & 30.6926           & 0.923854            & 0.153942             \\
                                & room     & 23.3253   & 31.4718           & 0.922728            & 0.2767               \\
                                & stump    & 88.6841   & 26.3107           & 0.756331            & 0.297154             \\
                                & treehill & 67.3652   & 22.1591           & 0.623254            & 0.40638              \\
                                & \textbf{AVG} & 57.1948 & 27.0237           & 0.802110            & 0.278507             \\
                                \hline
    \multirow{10}{*}{Small}     & bicycle  & 83.713    & 24.4335           & 0.733459            & 0.292415             \\
                                & bonsai   & 18.6092   & 31.7454           & 0.939945            & 0.246396             \\
                                & counter  & 18.6033   & 28.7987           & 0.906964            & 0.25618              \\
                                & flowers  & 55.762    & 21.2031           & 0.577685            & 0.401654             \\
                                & garden   & 85.358    & 26.1642           & 0.828               & 0.185808             \\
                                & kitchen  & 29.69     & 30.6111           & 0.922741            & 0.155509             \\
                                & room     & 20.4055   & 31.4142           & 0.922475            & 0.27728              \\
                                & stump    & 77.3974   & 26.2979           & 0.755284            & 0.29914              \\
                                & treehill & 58.4913   & 22.153            & 0.622094            & 0.407919             \\
                                & \textbf{AVG} & 49.7811 & 26.9801           & 0.800961            & 0.280256             \\
                                \hline
    \multirow{10}{*}{Very Small} & bicycle & 71.4395   & 24.4202           & 0.732045            & 0.294274             \\
                                & bonsai   & 15.7028   & 31.6188           & 0.938923            & 0.248307             \\
                                & counter  & 15.8929   & 28.7175           & 0.905882            & 0.258231             \\
                                & flowers  & 47.594    & 21.1754           & 0.574814            & 0.404517             \\
                                & garden   & 71.246    & 26.0987           & 0.823746            & 0.191957             \\
                                & kitchen  & 24.8944   & 30.4957           & 0.921001            & 0.157975             \\
                                & room     & 17.3102   & 31.3587           & 0.921477            & 0.279093             \\
                                & stump    & 65.737    & 26.2812           & 0.754719            & 0.30025              \\
                                & treehill & 49.5787   & 22.1406           & 0.620585            & 0.411527             \\
                                & \textbf{AVG} & 42.1551 & 26.9230           & 0.799244            & 0.282904             \\
                                \hline
    \multirow{10}{*}{Tiny}      & bicycle  & 62.1525   & 24.3973           & 0.729589            & 0.297164             \\
                                & bonsai   & 13.5061   & 31.5318           & 0.937773            & 0.25025              \\
                                & counter  & 13.5964   & 28.6485           & 0.904086            & 0.260673             \\
                                & flowers  & 40.5834   & 21.1358           & 0.572934            & 0.406518             \\
                                & garden   & 61.0334   & 25.9982           & 0.81838             & 0.199791             \\
                                & kitchen  & 20.7034   & 30.3032           & 0.917772            & 0.162754             \\
                                & room     & 15.2096   & 31.281            & 0.91991             & 0.281948             \\
                                & stump    & 56.3407   & 26.2594           & 0.753323            & 0.302683             \\
                                & treehill & 42.1912   & 22.1279           & 0.618904            & 0.413925            \\
                                & \textbf{AVG} & 36.1463 & 26.8537           & 0.796964            & 0.286190             \\
                        \hline
    \end{tabular}%
    }
    \endgroup
\end{table}

\begin{table}[t]
    \centering
    \caption{\textbf{Detailed results on Deep Blending.} We report the size and rendering quality of MesonGS++ under five target-size profiles.}
    \label{tab:supp_deep_blending}
    \resizebox{\linewidth}{!}{%
    \begin{tabular}{cccccc}
        \hline
    Type                        & Scene     & Size (MB) & PSNR ($\uparrow$) & SSIM ($\uparrow$) & LPIPS ($\downarrow$) \\
    \hline
    \multirow{3}{*}{Large}      & drjohnson & 61.7229   & 29.0222           & 0.898561            & 0.318163             \\
                                & playroom  & 49.1817   & 30.2752           & 0.906758            & 0.306569             \\
                                & \textbf{AVG} & 55.4523 & 29.6487           & 0.902660            & 0.312366             \\
                                \hline
    \multirow{3}{*}{Medium}     & drjohnson & 54.1369   & 29.0129           & 0.898347            & 0.318926             \\
                                & playroom  & 42.7007   & 30.266            & 0.906512            & 0.307227             \\
                                & \textbf{AVG} & 48.4188 & 29.6394           & 0.902429            & 0.313077             \\
                                \hline
    \multirow{3}{*}{Small}      & drjohnson & 47.4457   & 28.9912           & 0.897917            & 0.32024              \\
                                & playroom  & 37.0633   & 30.2511           & 0.906299            & 0.308087             \\
                                & \textbf{AVG} & 42.2545 & 29.6212           & 0.902108            & 0.314164             \\
                                \hline
    \multirow{3}{*}{Very Small} & drjohnson & 40.1182   & 28.9786           & 0.898112            & 0.320289             \\
                                & playroom  & 31.6761   & 30.2377           & 0.905984            & 0.308999             \\
                                & \textbf{AVG} & 35.8971 & 29.6082           & 0.902048            & 0.314644             \\
                                \hline
    \multirow{3}{*}{Tiny}       & drjohnson & 34.8362   & 28.956            & 0.897644            & 0.321631             \\
                                & playroom  & 27.5852   & 30.2176           & 0.90557             & 0.309945           \\
                                & \textbf{AVG} & 31.2107 & 29.5868           & 0.901607            & 0.315788             \\
                                \hline  
    \end{tabular}%
    }
    \end{table}

\begin{table}[t]
    \centering
    \caption{\textbf{Detailed results on Tank\&Temples.} We report the size and rendering quality of MesonGS++ under five target-size profiles.}
    \label{tab:supp_tandt}
    \resizebox{\linewidth}{!}{%
    \begin{tabular}{cccccc}
        \hline
    Type                        & Scene & Size (MB) & PSNR ($\uparrow$) & SSIM ($\uparrow$) & LPIPS ($\downarrow$) \\
    \hline
    \multirow{3}{*}{Large}      & train & 20.0593   & 21.78             & 0.804435            & 0.260408             \\
                                & truck & 42.1151   & 24.9694           & 0.873929            & 0.18465              \\
                                & \textbf{AVG} & 31.0872 & 23.3747           & 0.839182            & 0.222529             \\
                                \hline
    \multirow{3}{*}{Medium}     & train & 17.4636   & 21.7679           & 0.803621            & 0.261529             \\
                                & truck & 37.0383   & 24.9503           & 0.872773            & 0.186297             \\
                                & \textbf{AVG} & 27.2509 & 23.3591           & 0.838197            & 0.223913             \\
                                \hline
    \multirow{3}{*}{Small}      & train & 15.0683   & 21.7446           & 0.802214            & 0.263405             \\
                                & truck & 32.4929   & 24.901            & 0.872069            & 0.186955             \\
                                & \textbf{AVG} & 23.7806 & 23.3228           & 0.837141            & 0.225180             \\
                                \hline
    \multirow{3}{*}{Very Small} & train & 12.7776   & 21.6899           & 0.798386            & 0.267648             \\
                                & truck & 28.0126   & 24.8752           & 0.870751            & 0.188907             \\
                                & \textbf{AVG} & 20.3951 & 23.2825           & 0.834568            & 0.228278             \\
                                \hline
    \multirow{3}{*}{Tiny}       & train & 10.8851   & 21.6609           & 0.796549            & 0.270259             \\
                                & truck & 24.7253   & 24.8336           & 0.868258            & 0.191999            \\
                                & \textbf{AVG} & 17.8052 & 23.2472           & 0.832404            & 0.231129             \\
                                \hline
    \end{tabular}%
    }
\end{table}

\section{Definition of the ILP problem}

An \emph{Integer Linear Programming (ILP)} problem is a mathematical optimization model defined as:
\begin{equation*}
\begin{aligned}
& \underset{\mathbf{x}}{\text{minimize}}
& & \mathbf{c}^\top \mathbf{x} \\
& \text{subject to}
& & A\mathbf{x} \leq \mathbf{b}, \\
&&& \mathbf{x} \geq \mathbf{0}, \\
&&& \mathbf{x} \in \mathbb{Z}^n,
\end{aligned}
\end{equation*}
where \(\mathbf{x}\) is the vector of integer decision variables, \(\mathbf{c} \in \mathbb{R}^n\) defines the objective coefficients, \(A \in \mathbb{R}^{m \times n}\) is the constraint matrix, and \(\mathbf{b} \in \mathbb{R}^m\) represents the right-hand side coefficients. The notation \(\leq\) applies component-wise to vectors.

\section{Definition of the 0-1 ILP}
A \emph{0-1 Integer Linear Programming (0-1 ILP)} problem extends the ILP formulation with additional binary constraints, defined as:
\begin{equation*}
\begin{aligned}
& \underset{\mathbf{x}}{\text{maximize}}
& & \mathbf{c}^\top \mathbf{x} \\
& \text{subject to}
& & A\mathbf{x} \leq \mathbf{b}, \\
&&& \mathbf{x} \geq \mathbf{0}, \\
&&& \mathbf{x} \in \{0,1\}^n,
\end{aligned}
\end{equation*}
where all components must maintain \emph{linearity} in both the objective function and constraints. Specifically:
\begin{itemize}
    \item The objective function $\mathbf{c}^\top \mathbf{x}$ is linear in decision variables $\mathbf{x}$;
    \item All inequality constraints $A\mathbf{x} \leq \mathbf{b}$ are \textbf{linear combinations} of variables;
    \item Decision variables are restricted to binary values: $x_i \in \{0,1\},\ \forall i \in \{1,...,n\}$.
\end{itemize}

The matrix-vector notation $A\mathbf{x} \leq \mathbf{b}$ represents a system of $m$ linear inequalities, and both the objective and constraints must contain only first-order terms of variables without any products between variables or nonlinear functions.

\IEEEpubidadjcol

\section{Definition of the MINLP}
Mixed-Integer Nonlinear Programming (MINLP) refers to a class of optimization problems that involve both continuous and discrete (typically integer) decision variables, and include nonlinearities in the objective function and/or the constraints. Formally, a general MINLP can be written as:

\begin{align*}
\min_{x, y} \quad & f(x, y) \\
\text{s.t.} \quad & g_i(x, y) \leq 0, \quad i = 1, \ldots, m \\
& h_j(x, y) = 0, \quad j = 1, \ldots, p \\
& x \in \mathbb{R}^n, \quad y \in \mathbb{Z}^k
\end{align*}

Here:
\begin{itemize}
    \item $x \in \mathbb{R}^n$ are the continuous variables,
    \item $y \in \mathbb{Z}^k$ are the integer (often binary or bounded) variables,
    \item $f(x, y)$ is the nonlinear objective function to be minimized (or maximized),
    \item $g_i(x, y)$ and $h_j(x, y)$ are the nonlinear inequality and equality constraints, respectively.
\end{itemize}

MINLP combines the complexity of:
\begin{itemize}
    \item \textbf{Nonlinear Programming (NLP)}, which deals with nonlinear relationships,
    \item \textbf{Mixed-Integer Programming (MIP)}, which deals with linear problems involving integer variables and continuous variables.
\end{itemize}

Due to the presence of integer variables and nonlinearities, MINLP problems are generally non-convex and computationally hard to solve. Exact methods (such as branch-and-bound, outer approximation, and generalized Benders decomposition) and heuristic/metaheuristic algorithms (such as genetic algorithms, simulated annealing, etc.) are often employed depending on the problem structure and required solution quality.

\end{document}